%% file: cinematographic-drones.tex
\documentclass[tog]{acmsiggraph}

\graphicspath{{figures/}}
\usepackage{amsmath,amssymb,amsbsy}
\usepackage{algorithm}
\usepackage{algorithmic}
\usepackage{stmaryrd}
\usepackage{subcaption}

\newcommand{\I}{\mathcal{I}}
\newcommand{\isep}{\mathrel{{.}\,{.}}\nobreak}

\newcommand{\vect}[1]{\boldsymbol{\mathbf{#1}}}
\newcommand{\norm}[1]{\left\lVert#1\right\rVert}

\newcommand{\degree}[0]{^{\circ}}

\newcommand{\ie}{\emph{i.e.}~}
\newcommand{\eg}{\emph{e.g.}~}
\newcommand{\etal}{\emph{et~al.}~}
\newcommand{\wrt}{\textsl{w.r.t.}~}
\newcommand{\NB}{\textsl{NB:}~}

\newcommand{\resp}{respectively~}
 
\usepackage[usenames, dvipsnames]{color}

\title{Directing Cinematographic Drones}

\author{
Quentin Galvane\\Technicolor, Rennes, France
\and Christophe Lino, Marc Christie\\Inria, Rennes, France
\and Julien Fleureau, Fabien Servant, François-Louis Tariolle, Philippe Guillotel\\Technicolor, Rennes, France
}
\pdfauthor{Q. Galvane, C. Lino, M. Christie, J. Fleureau, F. Servant, F-L. Tariolle, P. Guillotel}

\TOGonlineid{0288}


\setcopyright{acmcopyright}

\copyrightyear{2017}

\TOGvolume{0}
\TOGnumber{0}

\hypersetup{draft}

\begin{document}


 \teaser{
	\centering
	\includegraphics[width=\linewidth]{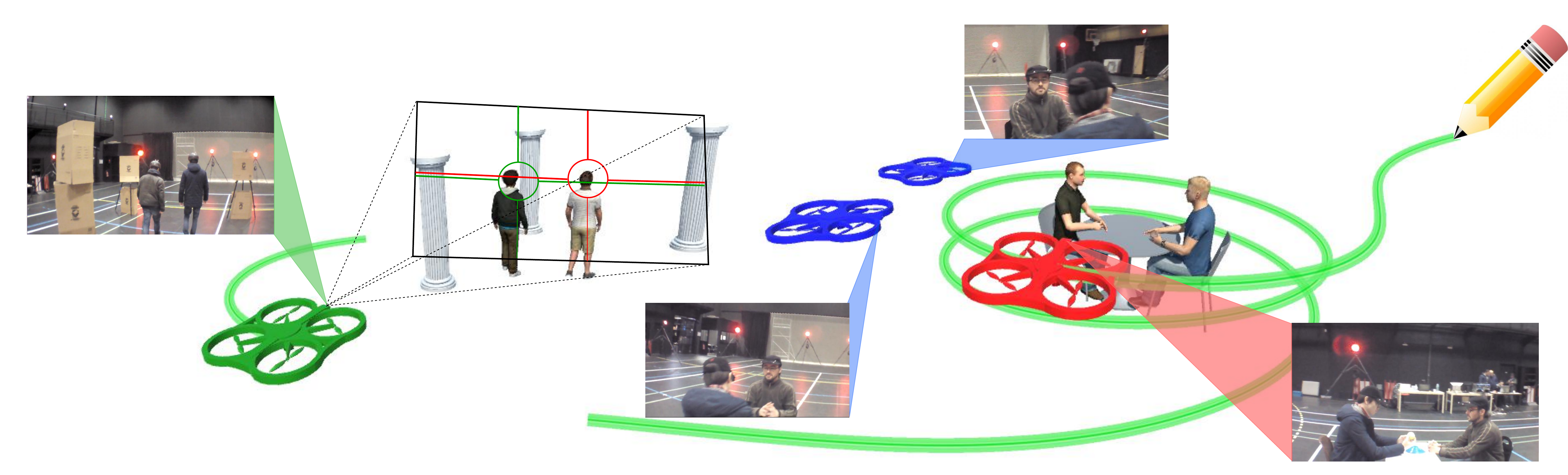}
	\caption{We introduce a set of high-level tools for filming dynamic targets with quadrotor drones. We first propose a specific camera parameter space (the Drone Toric space) together with on-screen viewpoint manipulators compatible with the physical constraints of a drone. We then propose a real-time path planning approach in dynamic environments which ensures both cinematographic properties in viewpoints along the path and  feasibility of the path by a quadrotor drone (see green quadrotor). We also present a sketching tool that generates feasible trajectories from hand drawn input paths (see red quadrotor). Finally we propose to coordinate positions and motions of multiple drones around the dynamic targets to ensure the coverture of cinematographic distinct viewpoints (see blue quadrotors).}
	\label{fig:teaser}
 }

\maketitle

\begin{abstract}


Quadrotor drones equipped with high quality cameras have rapidely raised as novel, cheap and stable devices for filmmakers. While professional drone pilots can create aesthetically pleasing videos in short time, the smooth -- and cinematographic -- control of a camera drone remains challenging for most users, despite recent tools that either automate part of the process or enable the manual design of waypoints to create drone trajectories. 

This paper proposes to move a step further towards more accessible cinematographic drones by designing techniques to automatically or interactively plan quadrotor drone motions in 3D dynamic environments that satisfy both cinematographic and physical quadrotor constraints. 
We first propose the design of a \emph{Drone Toric Space} as a dedicated camera parameter space with embedded constraints and derive some intuitive on-screen viewpoint manipulators.
Second, we propose a specific path planning technique which ensures both that cinematographic properties can be enforced along the path, and that the path is physically feasible by a quadrotor drone. At last, we build on the Drone Toric Space and the specific path planning technique to coordinate the motion of multiple drones around dynamic targets.
A number of results then demonstrate the interactive and automated capacities of our approaches on a number of use-cases.

\end{abstract}

%
%


%
%


\keywordlist

\conceptlist


\section{Introduction}
\input{introduction.tex}

\section{Related Work}
\label{sec:related-work}

\input{related-work.tex}

\section{Overview}
\label{sec:overview}
\input{overview.tex}

\section{DTS: a parametric space for drone control}

\label{sec:safe-toric-space}


\input{safe-toric-space.tex}

\subsection{Ensuring feasible camera orientations}
\label{sec:feasible-orientation}
\input{feasible-orientation.tex}

\section{Through-the-lens control of drones}
\label{sec:interactive-control}
\input{interactive-control.tex}


\section{Computing feasible drone trajectories}
\label{sec:planning-paths}
\input{planning-paths.tex}

\section{Coordinating cinematographic drones}
\label{sec:coordinating-drones}
\input{coordinating-drones.tex}

\section{Evaluation and Discussion}
\label{sec:results}
\input{results.tex}

\section{Conclusion}
\label{sec:conclusion}
\input{conclusion.tex}

\bibliographystyle{acmsiggraph}
\bibliography{cinematographic-drones}


\end{document}

%% file: introduction.tex

With the advent of stable and powerful quadrotors, coupled with high quality camera lenses mounted on controllable gimbals, quadrotors are becoming new cinematographic devices in the toolbox of both professional and amateur filmmakers. However, mastering the control of such devices to create desired sequences in possibly evolving environments requires a significant amount of time and practice. Indeed, synchronizing the quadrotor and camera motions while ensuring that the drone is in a safe position and satisfies desired visual properties remains challenging. Professional film crews actually rely on two operators who coordinate their actions: a pilot focuses on the drone motion and a cinematographer focuses on the camera orientation. 

To ease this process, consumer drone quadrotors have been proposing features such as \emph{follow-me}, in which the drone automatically follows and frames a moving target (using a GPS or vision-based tracking algorithms). More evolved approaches propose to design virtual trajectories in 3D environments, ensuring their feasibility before executing them in the real world~\cite{roberts2016,joubert2015interactive,Gebhardt:2016:AOP:2858036.2858353}. While such approaches lead to useful prototyping tools, they do not consider moving targets, and are therefore limited to static or close to static guided tour scenes. Collision constraints are not considered (except for~\cite{Gebhardt:2016:AOP:2858036.2858353}), and avoidance trajectories must be designed manually. In addition, little control of the visual properties is provided, such as maintaining a framing, distance or camera angle on a target.

In contrast, the computer graphics community has been focusing on how to enforce visual properties related to framing, distance or orientation of a virtual camera \wrt dynamic targets, and also enforce such properties over camera paths in virtual environments. Automated viewpoint computation tools can (i) solve such problems in real-time~\cite{ranon2014improving}, (ii) perform transitions between viewpoints by interpolating cinematographic properties~\cite{lino2015intuitive} and (iii) automatically edit sequences from multiple viewpoints~\cite{galvane2015editing}. The transposition of such techniques to quadrotors re

Recent approaches have been considering the application of such cinematographic visual properties to drones through mecanisms to ensure their satisfaction in static scenes~\cite{joubert2016cinematographer} and mechanims to follow predesigned camera paths using local avoidance techniques \cite{Naegeli2017}. 

There however are no fully dynamic planning techniques that can interactively plan multiple drone trajectories in relation to  dynamic targets. More importantly, the design of a cinematographic drone system with the ability to enforce visual properties on dynamic targets, requires to address a number of challenges. First, the formalization of cinematographic film principles need to be adapted with quadrotor drone constraints (limited view angles, path feasibility). Second, safety must also be ensured at any time during interactive and automated control, which in turn requires the adaptation of intuitive viewpoint manipulation tools (\cite{lino2015intuitive}). Third, the coordination of multiple drones requires to maintain and dynamically reposition cameras to ensure complementary cinematographic viewpoints around moving targets. 




We address these three challenges by proposing:
\begin{itemize}
\item a dedicated model for target-based drone placement -- the \emph{Drone Toric Space} --, which ensures both the feasibility of drone positions around targets and safety constraints with regards to these targets;
\item an interactive drone manipulation tool, which offers cinematographic and through-the-lens controls around targets;
\item dedicated real-time path planning techniques for dynamic environments, which enable the creation of cinematographic trajectories, further optimized to ensure their feasibility by quadrotor drones;
\item a coordination technique, to orchestrate the placement of multiple drones around dynamic targets, using min-conflict optimization techniques. 
\end{itemize}

To the best of our knowledge, this is the first system to provide both interactive and automated cinematographic control on one or multiple quadrotor drones for the specific task of framing targets in dynamic environments. This enables to envision smarter design tools for the creation of cinematographic sequences, where users would essentially focus on the aesthetic choices. This also opens perspectives towards controlling autonomous groups of drones with cinematographic behaviors which would enable the prototyping of film sequences or the shooting of documentaries.


%% file: related-work.tex
\paragraph{Automated camera control in virtual environments}

The problem of controlling a camera in a virtual 3D environment has been addressed by a wide range of techniques and is strongly guided by the type of tasks to perform and the target application. An overview is presented in~\cite{christie2009camera,christie2008camera} gathering automated, reactive and interative approaches to virtual camera control, including specific techniques for planning paths, managing occlusions and modeling high-level communicative goals. 

 We here restrict our overview to techniques closely related to our approach. The automated computation of viewpoints has first been addressed by Blinn~\cite{blinn1988looking} who proposed an efficient iterative technique to compute the position and orientation of a camera from the specification of on-screen positions and visual properties. The problem has been expressed in a more general framework where visual properties in the image space (position an orientation of targets) are expressed as constraints on the degrees of freedom of the camera, and been solved through a range of techniques including stochastic, regular sampling or quadratic programming \cite{drucker1994intelligent,bares2000model,ranon2014improving}. Recently, a different camera representation has been proposed, the Toric Space, that simplifies the expression and solving of viewpoint computation problems~\cite{lino2015intuitive}.
 
The computation of camera paths imposes challenges such as collision with complex 3D environments, visibility of multiple targets, and also smoothness over the trajectory. In ~\cite{salomon2003interactive}, the authors present an approach for interactive navigation in complex 3D synthetic environments using path planning. A collision-free and constrained path between two user specified locations can be computed on demand by relying on a prior construction of a global roadmap of the environment using randomized motion planning and graph search techniques (here an IDA* depth-first search). Smoothness is enforced by simply cutting corners along the path. More recently, Oskam etal.  presented an approach that generates camera paths and enforces visibility of a target along the path when possible~\cite{oskam2009visibility}. The process relies on a prior sphere-sampling stage in which the visibility between every pair of spheres is precomputed. Adjacent spheres are used to construct a graph then traversed by using an A* planner. The cost on the arcs is a combination of distance and visibility. A specific smoothing process is applied which maximises the visibility along the trajectory


To the best of our knowledge, no approach has been coupling path planning techniques to maintain or to interpolate visual properties in the context of cinematographic drones. Furthermore, specific constraints on the continuity of the generated paths must be set to ensure the feasibility by a drone, which requires a strong adaptation of existing techniques.

\paragraph{Trajectory planning for drones}

Adding to UAVs (unmanned aerial vehicles) the capacity to take photos and shoot sequences has triggered the development of a number of techniques. Applications range from automated surveillance tasks to area coverage, scanning of unknown environments, or capture of aesthetic shots of buildings, landscapes and characters. All approaches have in common the computation of trajectories that have to obey the physical characteristics of the UAV motion. 

For scanning unknown environments, different strategies have been applied: Dunkley \etal.~\shortcite{dunkley14iros} perform autonomous hovering with a quadrotor drone using a visual-inertial SLAM (simultaneous localization and mapping) system, Nuske \etal.~\shortcite{nuske2015autonomous} propose a system that detects and maps a specific visual feature (river), and plans paths around three-dimensional (3D) obstacles (such as overhanging tree branches) only with onboard sensing and no GPS nor prior map. 

In approaches where the environment is known, research has also been focusing on the generation of optimal safe trajectories while satisfying constraints on velocities and accelerations bounds for one drone~\cite{mellinger2011minimum} or for a set of drones, each with a specific goal state~\cite{turpin2013goal}. Deits \etal~\shortcite{deits2015efficient} present an approach to the design of smooth trajectories for quadrotor UAVs, which are free of collisions with obstacles along their entire length. 

Multiple approaches have been addressing the problem of spatially coordinating entities~\cite{pereira2003formation}, to maintain specific spatial configurations for drones~\cite{schiano2016rigidity} or for drone collaborative tasks~\cite{mellinger2011minimum}, through centralized or decentralized systems.

  
The specific problem of assisting the design of drone trajectories for aesthetic aerial videography has received limited attention. Current approaches focus on the design of feasible trajectories that link user-defined viewpoints~\cite{joubert2015interactive,Gebhardt:2016:AOP:2858036.2858353,roberts2016}. The process consists in prototyping a trajectory in a 3D simulator before executing it automatically in the real environment. The virtual trajectory is designed by creating an ordered collection of look-from/look-at viewpoints (keyframes) manually positioned. In~\cite{joubert2015interactive}, the timing of the keyframes is also specified by the user. A specific $C^4$ continuous trajectory is then created between the keyframes ($C^4$ property ensures the path obeys the physical equations of motion). The resulting trajectory is then analyzed to detect the infeasible sections along the path (sections where the velocity or control force to be applied are too important), so the user can iteratively alter the keyframe timings. The work has later been extended to address this feasibility issue automatically~\cite{roberts2016}. The technique consists in performing a time-warping of the trajectory, altering the speed between the viewpoints without altering the trajectory nor the keyframes. 
 
Another quadrotor trajectory design tool has been proposed in~\cite{Gebhardt:2016:AOP:2858036.2858353}. The principle is similar to ~\cite{joubert2015interactive}: a camera path can be drawn and edited in a virtual environment, and then optimized to ensure its feasibility. However given the multiple constraints (including preventing collisions with the environment), the optimization process does not guarantee to respect the user inputs (a tradeoff between user inputs and conflicting constraints is performed). The tool offers an intuitive tool for novice users to create quadrotor based use-cases without requiring deep knowledge in either quadrotor control or the underlying constraints of the target domain.

Such approaches however do not account for more cinematographic properties on the viewpoints nor on the camera path. Fleureau \etal \shortcite{Fleureau2016} recently presented a tool to automatically maintain visual on-screen properties (orientation, composition) on moving targets, and automatically compute transitions between viewpoints with moving targets~\cite{wiced2016}. The approach relies on the Toric Space representation~\cite{lino2015intuitive} to efficiently express cinematographic properties (distance to target, angle on target, screen positions of targets) and perform interpolations in the Toric Space rather than in the Cartesian space to  maintain visual properties along the trajectory. 
Focusing on the following of predefined paths, Naegeli \etal \shortcite{Naegeli2017} proposed an online optimization scheme that computes drone control inputs to locally adjust the flight plan with a given time horizon and ensure its feasibility. Their solution allows the tracking and avoidance of dynamic targets.

In these approaches, while the generated trajectories satisfy the feasibility criteria, either the cameras are guided by users through keyframes and are limited to static scenes or they do not enable the autonomous planning of paths in dynamically changing environments. By contrast, the method we propose is fully automated, plans paths globally, works in a dynamic context and provides the foundations for collaborative motions of drones.

\paragraph{Viewpoint control in image space}

In computer graphics, different approaches have been proposing {\it Through-the-lens} camera control techniques, which are interactions that occur in the screen space to constrain the camera parameters~\cite{gleicher1992through,singh2009cubecam,lino2015intuitive}. The problem is generally expressed as a minimization between the user's specification and the current view properties. The techniques have been designed to control virtual cameras and to the best of our knowledge have not been applied to drone control (apart from straighforward forms of control such as look-from/look-at).

%% file: overview.tex
\begin{figure}[t!]
	\centering
	\includegraphics[width=\linewidth]{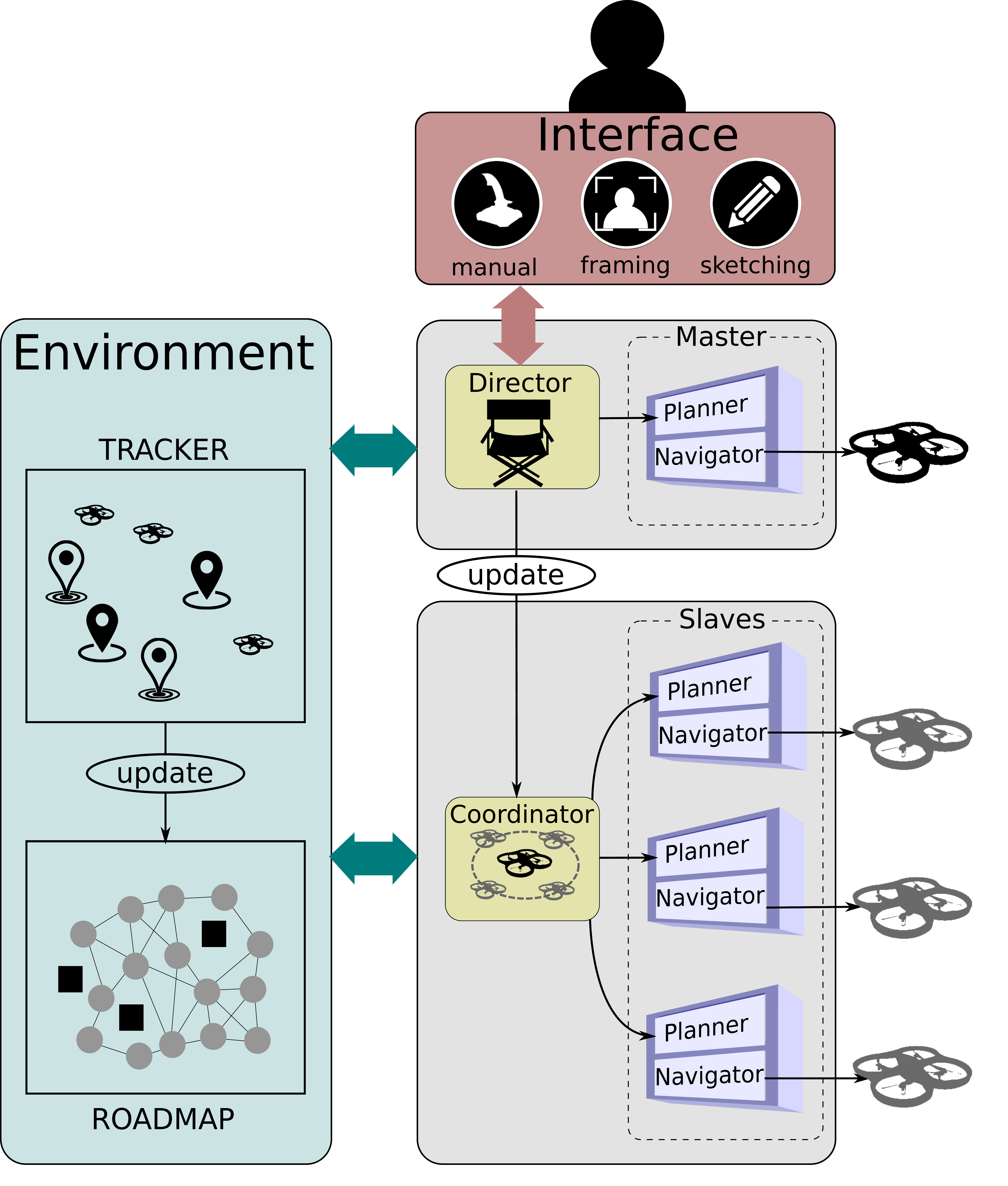}
	\caption{Overview of our drone system. Poses of drones and targets are tracked in real-time and exploited to update a roadmap which abstracts the real environment. Our path-planning process (planner) creates feasible trajectories in the roadmap, and relies on a navigator to follow the path.  A coordinator component orchestrates the motions of the remaing drones to create complementary shots. The users have high-level interactions on the master drone (perform framing and sketching).}
	\label{fig:overview}
\end{figure}



Our system is structured around two major components: the \emph{director} and the \emph{coordinator} (see  Figure~\ref{fig:overview}). The director component takes as input high-level user specifications such as a sketched trajectory to follow, or a framing to achieve in relation to targets. It then relies on a \emph{path-planner} component that offers different planning strategies (for each type of user interaction) and a \emph{navigator} component that computes the control commands to steer the drone along the computed path. The \emph{path planner} constantly checks the validity of the path (for collisions) and triggers its re-computation when necessary. 

The \emph{director} component controls the Master drone. The \emph{coordinator} component can then be used to orchestrate the motion of multiple drones (named slave drones) offering complementary views on the scene. Our system is connected to a localization system (the \emph{tracker}) which performs the real-time tracking of drones, targets and dynamic obstacles. This information is used to update a roadmap representation of the environment used by the planner.

To ensure that visual properties are enforced along the paths, the planning is performed in a dedicated space, the \emph{Drone Toric Space} (DTS). This space, presented in Section~\ref{sec:safe-toric-space}, is a re-parameterization of the Toric Space, a camera representation to express and manipulate viewpoints in computer animation. We show how this space can be exploited to propose through-the-lens manipulation techniques dedicated to quadrotor drones (Section~\ref{sec:interactive-control}), and how planning can be performed to avoid variations of on-screen properties (Section~\ref{sec:planning-paths}). We finally present how to orchestrate the motion of multiple drones to ensure cinematographic coverage of a 3D scene with dynamic targets by exploiting this Drone Toric Space (Section~\ref{sec:coordinating-drones}).

%% file: safe-toric-space.tex

In the following, we describe the Drone Toric Space, a representation dedicated to the manipulation and planning of cinematographic drone viewpoints. Our representation builds on an existing model, the Toric Space \cite{lino2015intuitive}, proposed for the control of virtual cameras. The Toric Space is an expressive camera model allowing to encode the visual properties of two filmed targets, and to reduce the complexity of viewpoint computations. The power of this model is that it directly encodes the screen positions of targets in the representation. Given two desired screen positions, the set of possible camera viewpoints is a 2-parametric surface (Toric surface) -- a spindle torus on which every point enforces the same angle $\alpha$ (computed from the screen positions) between the two targets and the camera. One can then parametrize the position and orientation of the camera onto this surface through two Euler angles, $\varphi$ and $\theta$, denoting its vertical and horizontal angle around targets respectively. In the Toric Space, any solution camera viewpoint can thus be described as a 3d vector $(\alpha,\varphi,\theta)$. More than just an extension, our \emph{Drone Toric Space} is designed in mind to also overcome some limitations of the Toric Space. Our Drone Toric Space accounts for (i) the safety of targets (\ie we enforce a safety distance $d_S$ to targets) and (ii) the constraints and limitations of quadrotor drones (\eg fixed or gimbal camera, limited tilt angle). 

In the following, we denote a drone configuration, for which the filming camera is the end-effector (see Figure~\ref{fig:drone-configuration}), as a 7d vector $\vect{q}(x,y,z,\rho,\gamma,\psi,\lambda)$ further decomposed as follows. The drone position is determined by a 3d vector $\vect{\omega}(x,y,z)$ in Cartesian space. Its orientation is determined by three Euler angles $(\rho,\gamma,\psi)$ representing its roll, pitch and yaw respectively. The camera's additional orientation is determined by the Euler angle $\lambda$ corresponding to the tilt of the gimbal holding the camera. We finally assume for the sake of simplicity that the drone and the filming camera position coincide. A \emph{target} is an object (typically a person) for which a number of visual properties should be enforced. A target is characterized by it's position and orientation $\vect{p}(x_p,y_p,z_p,\phi_p,\theta_p,\psi_p)$.


\subsection{Ensuring safe drone locations}

The Drone Toric Space first enforces collision avoidance with the drone's targets by construction. Assuming that a Toric surface is computed around two targets $A$ and $B$, some camera viewpoints may not enforce the safety distance (\eg viewpoints behind a target). We replace all these viewpoints (\ie drone positions forming a continuous surface) which are unsafe for a given target $T$ (either $A$ or $B$) by an alternative surface $\textsl{E}^T_S$. This new surface must be tangent to both the remaining Toric surface and a safety sphere of center $T$ and radius $d^S$. 

\begin{figure} [t!]
	\centering
	\begin{tabular}{c c}
		\includegraphics[height=5cm]{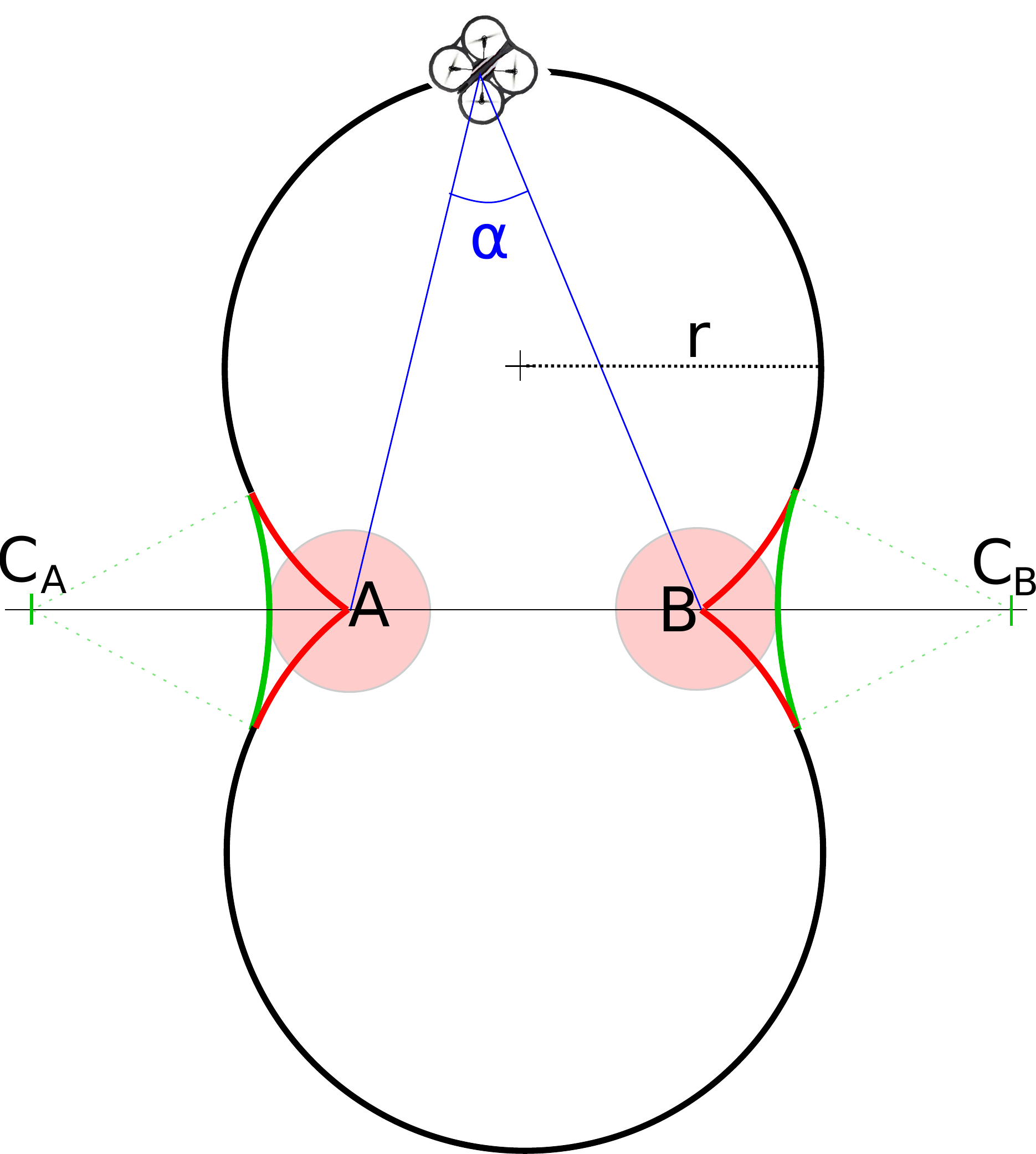} &
		\includegraphics[height=5cm]{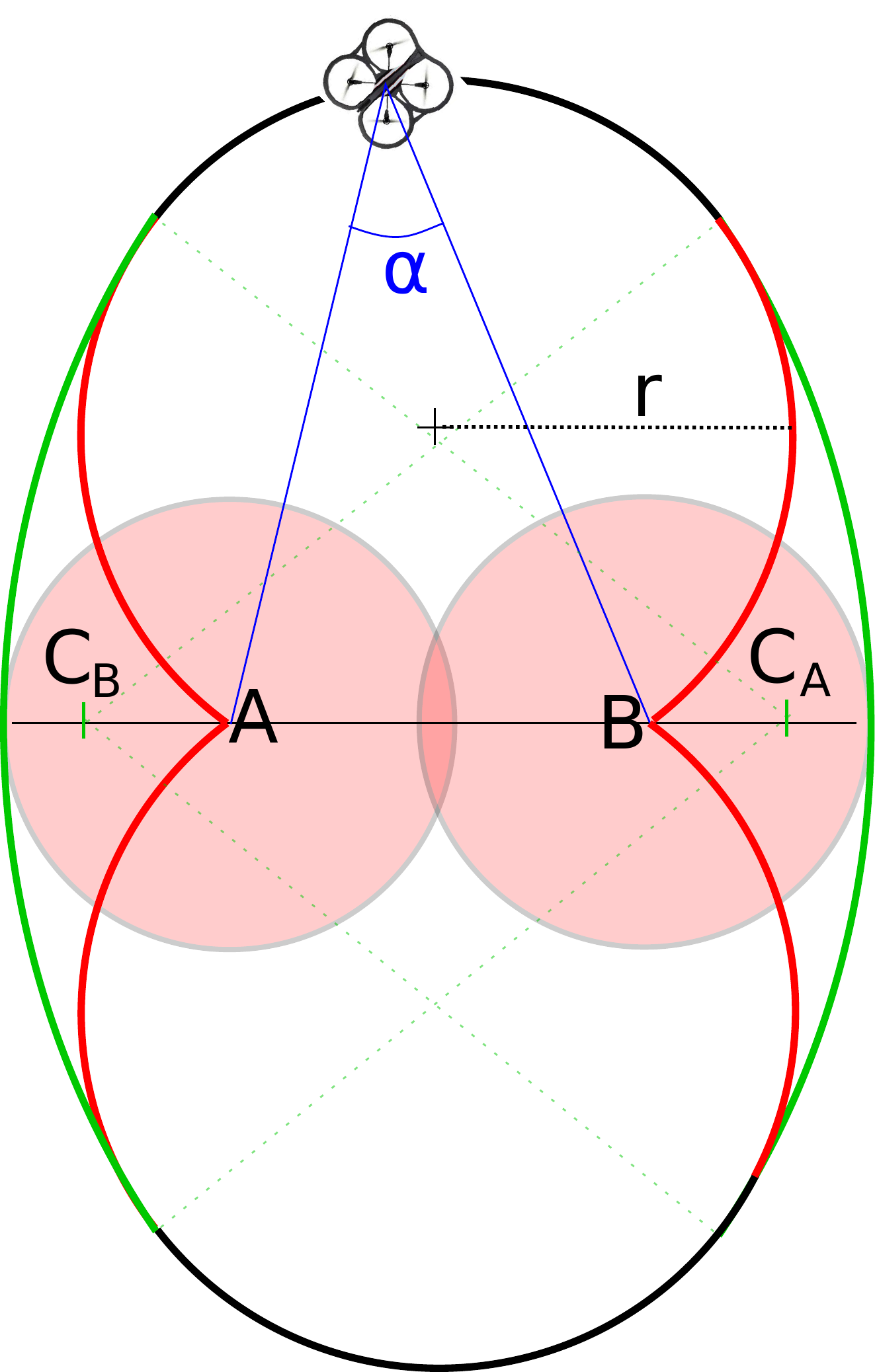} \\
		Type \#1 &
		Type \#2 \\
		\multicolumn{2}{c}{\includegraphics[height=5cm]{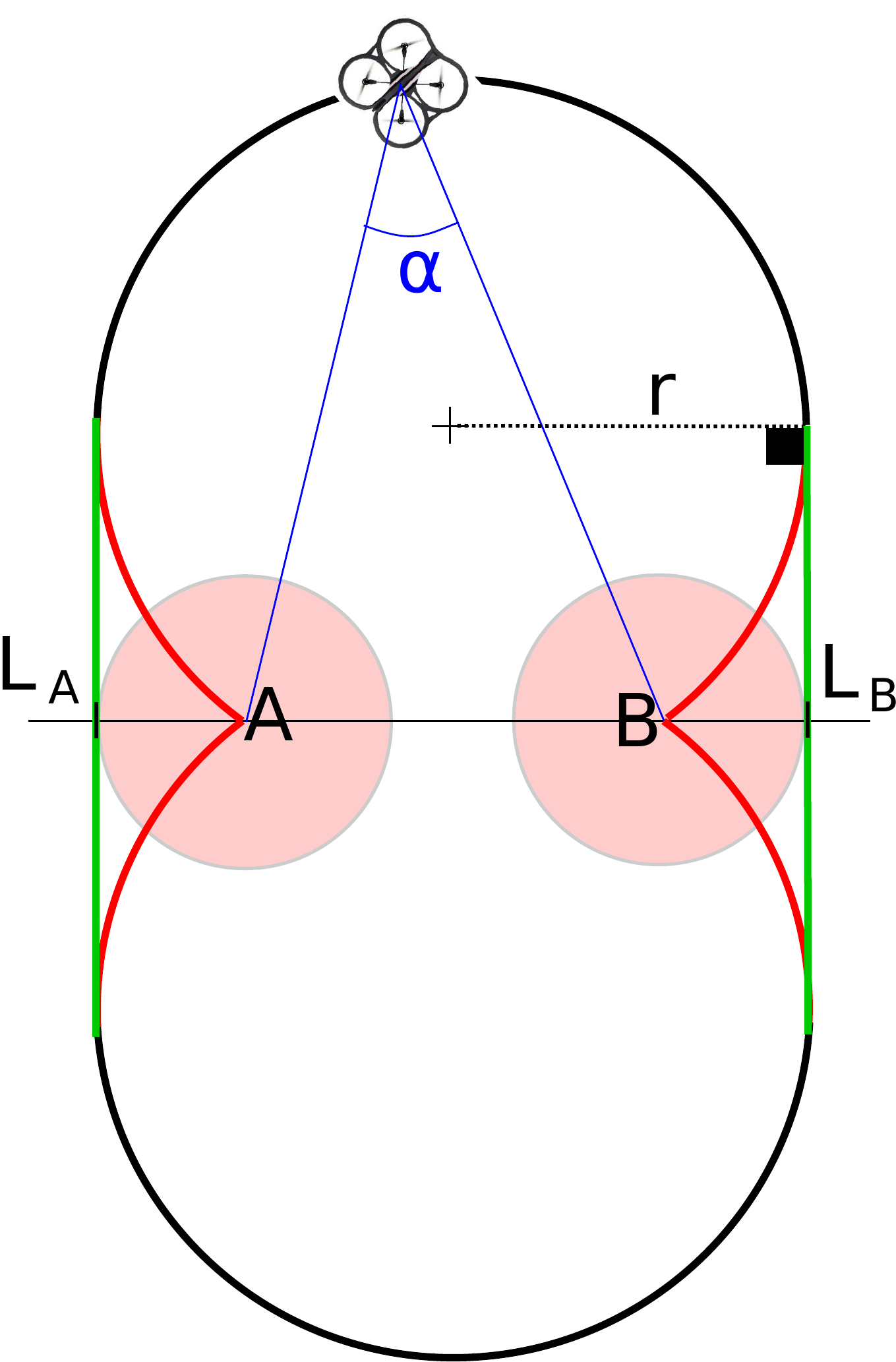}} \\
		\multicolumn{2}{c}{Type \#3} \\
	\end{tabular}
	\caption{ Red areas represent drone configurations which are unsafe for a target, \ie closer than a safety distance $d_S$. These areas are replaced by safe viewpoints (green surfaces) in the DTS model. This ensures that all drone viewpoints enforce the safety distance and that the Drone Toric surface (concatenation of black and green surfaces) is $C^1$ continuous. The type of green surface to use is strongly dependent on (i) the radius $r$ and the distance between the targets, with relation to (ii) the safety distance $d_S$ around targets.}
	\label{fig:safe-toric-space}
\end{figure}

In practice, we note $r$ the radius of the original Toric surface (figure~\ref{fig:safe-toric-space} shows a cross sections of a Drone Toric Space). Then, to determine the surface $\textsl{E}^A_S$, we rely on the 1-parametric line (of abscissa $x$) whose origin is target $A$ and whose directing vector points towards $B$. Depending on $d_S$, three types of Drone Toric surfaces can be distinguished:

{\bf Type \#1: $d^S < r-\frac{AB}{2}$.} $\textsl{E}^A_S$ belongs to a sphere, centered at a point $C_A$ of abscissa $x=-\frac{r^2-\delta^2}{2(\delta-r\sin\alpha)}$ (where $\delta=r-d^S$) and with radius $x-d^S$.

{\bf Type \#2: $d^S > r-\frac{AB}{2}$.} $\textsl{E}^A_S$ belongs to a sphere, centered at a point $C_A$ of abscissa $x=\frac{AB}{2}+\frac{\delta^2-r^2}{2(\delta-r\sin\alpha)}$ (where $\delta=d^S+AB-r$), and with radius $x+d^S$. 

{\bf Type \#3: $d^S = r-\frac{AB}{2}$.} $\textsl{E}^A_S$ belongs to a plane, orthogonal to the line $(AB)$, and passing through a point $L_A$ of abscissa $-d^S$. It corresponds to the limit between type \#1 and \#2 (\ie when $|x|$ tends toward $\infty$).

\NB surfaces of type \#1 are concave, while surfaces of type either \#2 or \#3 are convex (see figure \ref{fig:safe-toric-space}).

The Drone Toric surface is then parameterized through a pair of ratios $(\varphi,\theta)\in[-1;+1]^2$ representing the horizontal and vertical angles around targets. We have designed these ratios in a meaningful way:
practically \mbox{$\theta=0$} and \mbox{$\theta=\pm 1$} provide a view from behind $B$ and a view from behind $A$, respectively. Then, \mbox{$\varphi=0$}, \mbox{$\varphi=+1$} and \mbox{$\varphi=-1$} provide a view from the targets' height, from above and from below targets, respectively. 

\begin{figure} [t!]
	\centering
	\begin{tabular}{c}
	\includegraphics[height=5cm]{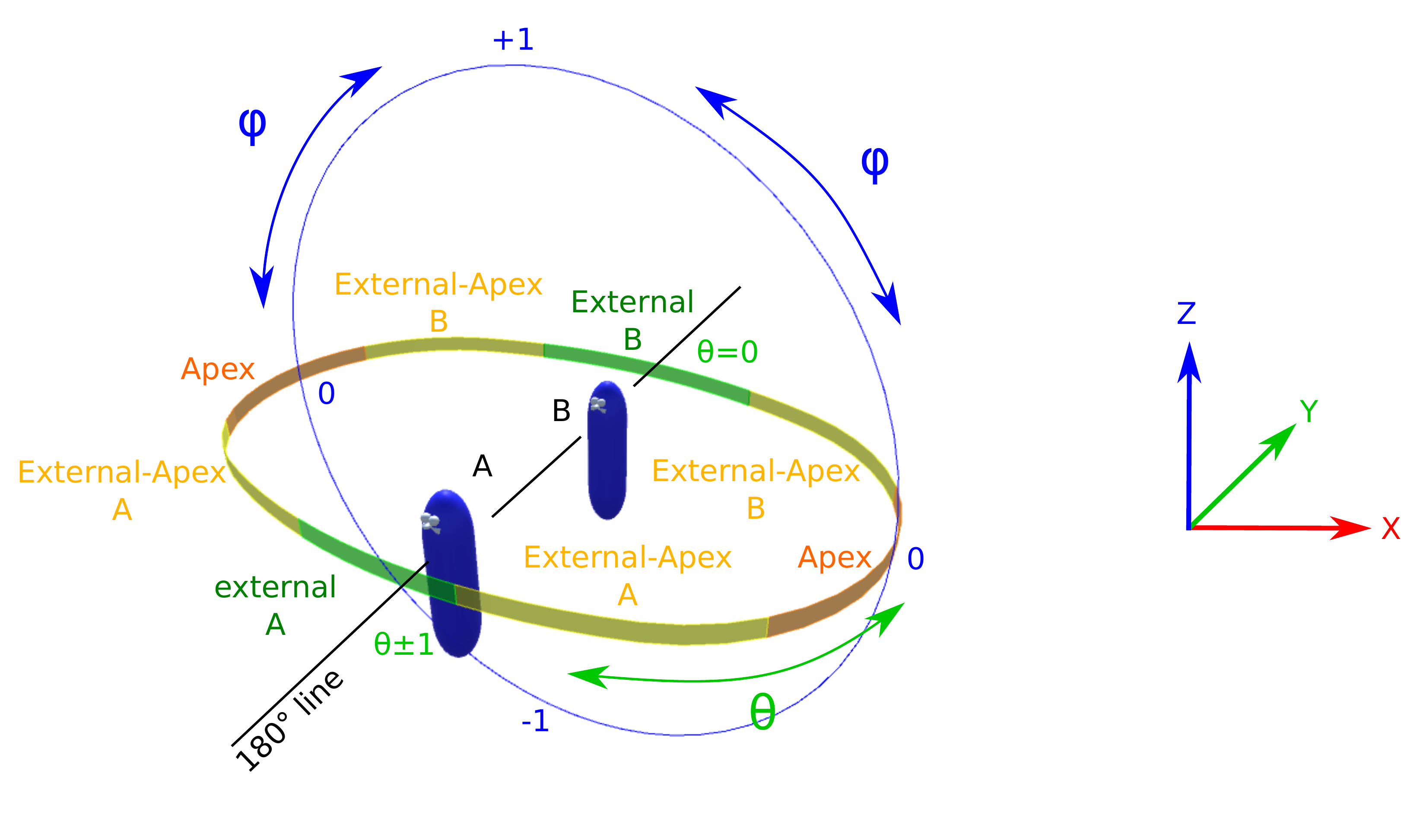} \\
	\includegraphics[height=5cm]{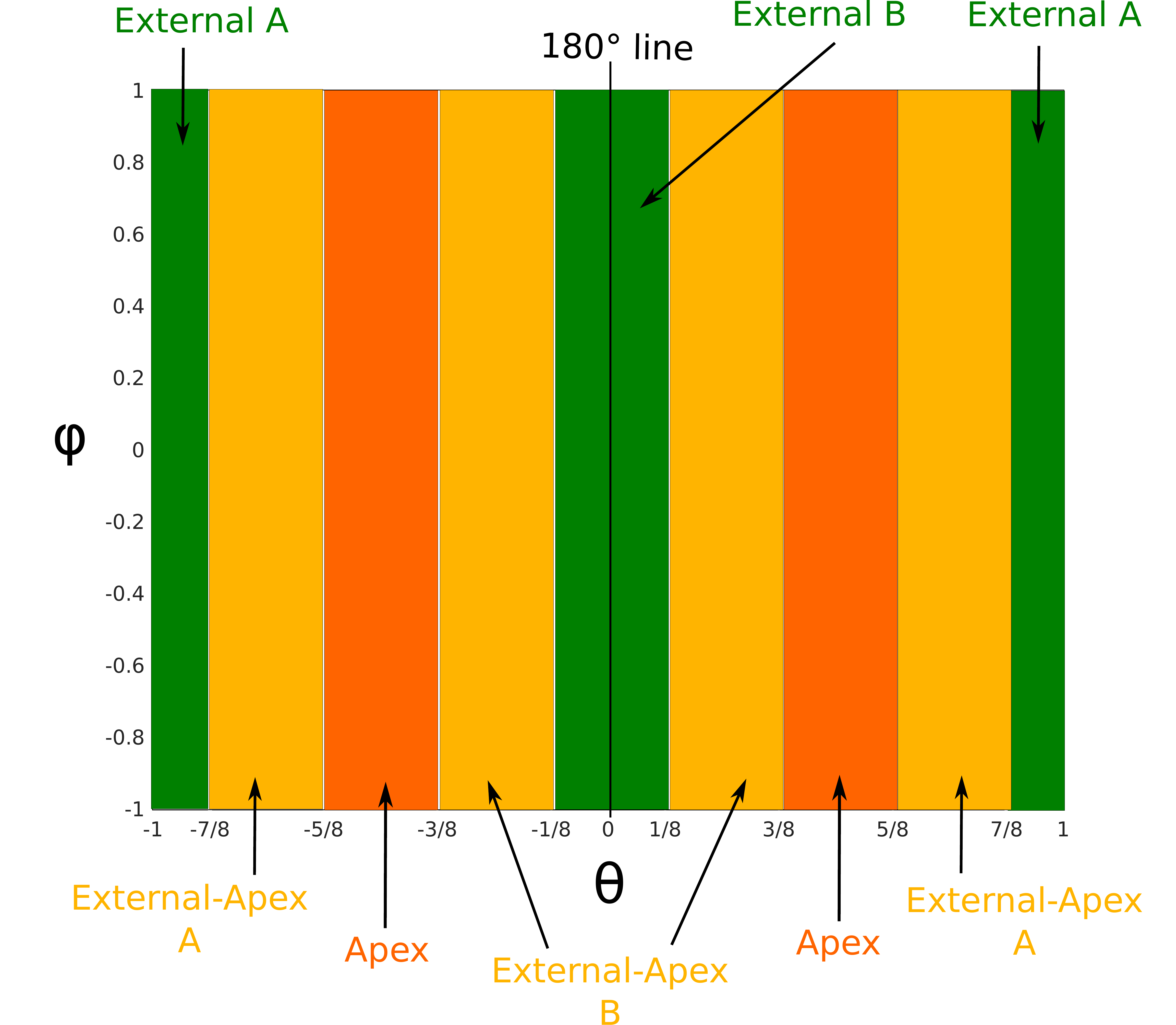}
	\end{tabular}
	\caption{Our Drone Toric Space parameterization features: specific camera regions are designed around two targets to ensure the safety of the targets. Classical horizontal view angles used in cinematography are cast into 3 semantic regions (external, external-apex, apex) modeled as intervals on parameter $\theta$. This ensures reversibility of interactions in manipulation tasks (\ie a reversed interaction will put the drone back in its initial position). Top: world space; Bottom: Drone Toric surface.}
	\label{fig:camera-regions}
\end{figure}

Further, following Arijon's triangle principle \cite{Arijon76}, the space of view angles and their allowed variation can be split into two main regions : \emph{external} views (where the camera should stay behind one target) and \emph{apex} views (where the camera should stay at equi-distance to targets). To allow a compromise between these two behaviors, we also define an intermediate region which we refer to as \emph{external-apex} views. In practice, we represent these characteristic camera regions through predefined intervals (of width 0.25) on the axis $\theta$ (see Figure \ref{fig:camera-regions}).

Note that, in the case of filming a single target, we can also build a Drone Toric surface, reduced to a sphere around the target, whose radius is never smaller than the safety distance. For the sake of simplicity, in this case we use the symbol $\alpha$ to refer to the sphere radius. Further, we accordingly parametrize this sphere through the same pair of ratios $(\varphi,\theta)$, where \mbox{$\theta=0$} and \mbox{$\theta=\pm 1$} provide a view from the front and from behind the target, respectively.

In the remaining, we will refer to such a re-parametrized surface as a \emph{Drone Toric surface} and to the continuous set of these surfaces defined around one or more targets as a \emph{Drone Toric Space}, parametrized through a triplet $(\alpha,\varphi,\theta)$.

%% file: feasible-orientation.tex
\begin{figure} [t!]
	\centering
	\includegraphics[width=0.5\linewidth]{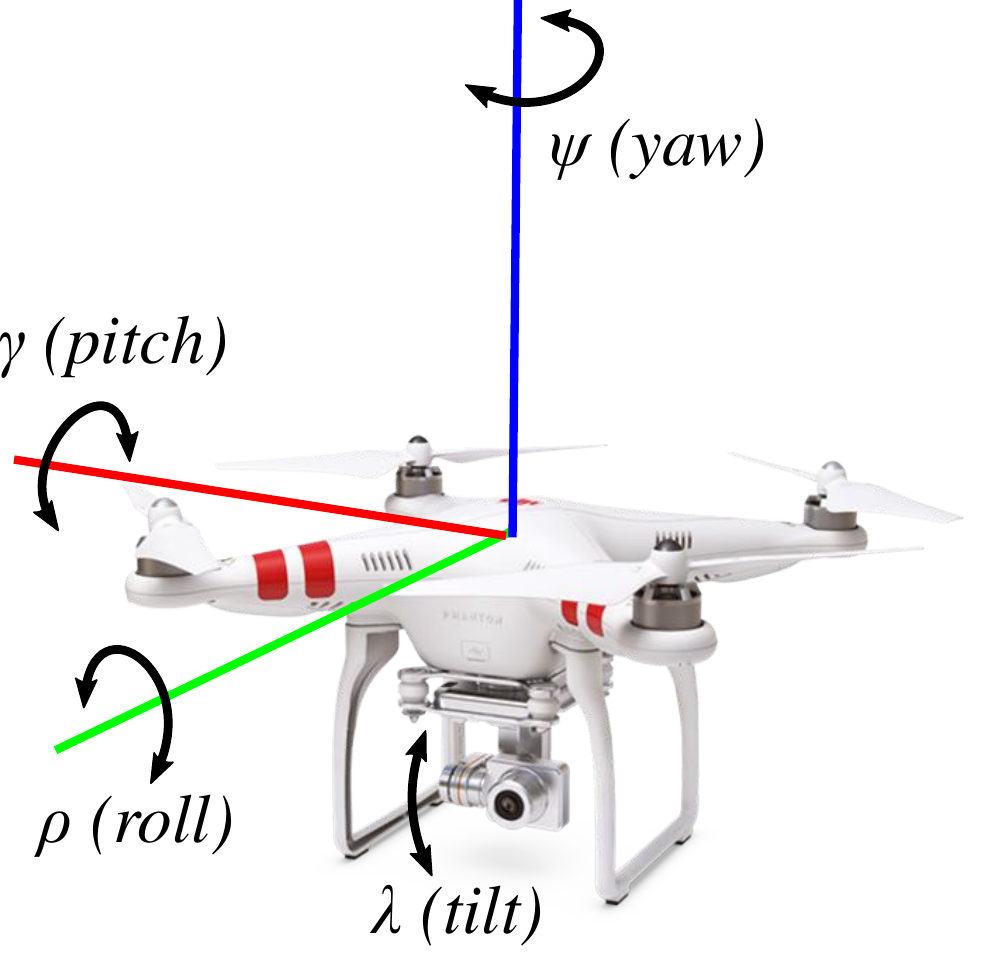} 
	\caption{Our drone's configuration (where the filming camera is considered as the end-effector) is determined by a 7d vector $\vect{q}(x,y,z,\rho,\gamma,\psi,\lambda)$, where the 3d vector $\vect{\omega}(x,y,z)$ represents the drone's location, the 3d vector $(\rho,\gamma,\psi)$ the drone's orientation and $\lambda$ the gimbal rotation. In our representation, we also assume that the camera position and the drone position coincide, as in practice the distance between them is negligible.}
	\label{fig:drone-configuration}
\end{figure}

Once a feasible drone location has been selected in our Drone Toric Space, we can compute a feasible camera orientation.
It is important to note that, assuming that the drone is set at a stationary location, the possible variations on the end-effector camera orientation will be limited to two degrees of freedom: its yaw ($\psi$) and tilt ($\lambda$) rotations. Indeed, as soon as either the drone's roll ($\rho$) or pitch ($\gamma$) is non null then the drone is set in motion (see Figure~\ref{fig:drone-configuration}). Further, to account for the physical limits of the gimbal rotation, $\lambda$ must be within an interval $\mathcal{I_\lambda}$ of feasible tilt rotations; in particular when no gimbal is used, $\mathcal{I_\lambda} = \left[ 0 \right]$. We here propose an iterative two-step algorithm that, given a stationary drone location and the desired screen positions of an arbitrary number of targets, computes a feasible drone orientation (its pseudo-code is given in Algorithm \ref{code:feasible-orientation}). 

In a first stage, we will assume that the initial camera orientation is feasible (\ie $\rho=0$, $\gamma=0$ and $\lambda \in \I_\lambda$); we explain how we compute this initial orientation later. From then on, we optimize the two degrees of freedom $\psi$ and $\lambda$ to best match the desired on-screen positioning of the filmed targets, while ensuring that the new orientation remains feasible. Practically, for a given camera orientation, we determine the error (horizontally or vertically) on their on-screen positioning. We express these errors as angular variations $\Delta \psi$ (yaw angle error) and $\Delta \lambda$ (tilt angle error) that we will apply on $\psi$ and $\lambda$. In our algorithm, we iterate on correcting each degree of freedom separately, while the other is left fixed, until both errors become lower than a predefined threshold $\epsilon$. 

We now detail how we compute $\Delta \psi$ and $\Delta \lambda$.
We here model the camera orientation as a Euclidian frame $(\vect{x_{c}},\vect{y_{c}},\vect{z_{c}})$ (\ie an east-north-up frame), centered at the camera's location $\omega$, and which can be determined from the 4d rotation vector $(\rho,\gamma,\psi,\lambda)$. Note that a key feature of a feasible camera orientation (whose roll is null) is that the east component $x_c$ is aligned with the horizon (\ie it must be orthogonal to the world up vector, which we refer to as $\vect{z_w}$). Thus, a variation of its yaw angle will correspond to a rotation around axis $\vect{z_w}$ and a variation of its tilt angle to a rotation around axis $\vect{x_c}$. We build on this feature to ensure the camera orientation remains feasible.

We assume that we want each target $T$ to be positioned at a 2d point $\vect{p_T}$ on the screen. Note that, in the world space, all 3d world points projecting at $p_T$ on the screen are located on a half-line (whose origin is the camera); in particular, we will refer to its directing vector as $\vect{v_T^d}$. We then compute the tilt and yaw errors for target $T$ (referred to as $\Delta \psi_T$ and $\Delta \lambda_T$). To do so, we rely on the difference between two vectors: (i) the desired direction $\vect{v_T^d}$ and (ii) the actual direction $\vect{v_T^a}$ from the camera to the target $T$. Practically,
\begin{equation*}
\Delta \psi_T = \measuredangle_{\vect{z_w}}\left[\Pi_{\vect{z_w}}(\vect{v_T^d}), \Pi_{\vect{z_w}}(\vect{v_T^a})\right]
\end{equation*}
where $\Pi_{\vect{n}}(\vect{v})$ is a projection operator, which projects vector $\vect{v}$ in the plane whose normal is $\vect{n}$; and $\measuredangle_{\vect{n}}\left[\vect{v_1}, \vect{v_2}\right]$ is the directed angle (in the plane whose normal is $\vect{n}$) between two vectors $\vect{v1}$ and $\vect{v_2}$.
We then compute $\Delta \psi$ as the average of all $\Delta \psi_T$; and apply the variation $\Delta \psi$ to $\psi$.
In a way similar,
\begin{equation*}
\Delta \lambda_T = \measuredangle_{\vect{x_c}}\left[\Pi_{\vect{x_c}}(\vect{v_T^d}), \Pi_{\vect{x_c}}(\vect{v_T^a})\right]
\end{equation*}
We compute $\Delta \lambda$ as the average of all $\Delta \lambda_T$, and apply the variation $\Delta \lambda$ to $\lambda$. To account for the range of feasible tilt values, we then clamp $\lambda$ to the interval $\mathcal{I_\lambda}$, and we clamp $\Delta \lambda$ to reflect the actual variation that was applied.
After each separate correction, we also recompute the frame $(\vect{x_{c}},\vect{y_{c}},\vect{z_{c}})$, as well as all vectors $\vect{v_T^d}$ to match the new camera orientation. 

\begin{algorithm}
\caption{Computation of a feasible orientation for $N$ targets}
\label{code:feasible-orientation}

\begin{algorithmic}[1]
\STATE // Initialize orientation
\STATE $(\rho_0, \gamma_0, \psi_0, \lambda_0)$ := computeInitialOrientation($\omega$, $targets$) 
\STATE clamp $\lambda_0$ to $\I_\lambda$
\STATE 
\STATE // Iterate on correcting $\psi_i$ and $\lambda_i$
\REPEAT
	\STATE // Correct yaw
	\FORALL {target $T$ in $targets$}
		\STATE $\Delta\psi_T$ := computeYawError($q$, $T$)
	\ENDFOR
	\STATE $\Delta\psi_i$ := $\frac{1}{N}\sum_T{\Delta \psi_T}$
	\STATE $\psi_i$ := $\psi_{i-1} + \Delta\psi_i$
	\STATE
	\STATE // Correct tilt
	\FORALL {target $T$ in $targets$}
		\STATE $\Delta\lambda_T$ := computeTiltError($q$, $T$)
	\ENDFOR
	\STATE $\Delta\lambda_i$ := $\frac{1}{N}\sum_T{\Delta \lambda_T}$
	\STATE $\lambda_i$ := $\lambda_{i-1} + \Delta\lambda_i$
	\STATE clamp $\lambda_i$ to $\I_\lambda$
	\STATE
	\STATE // clamp $\Delta\lambda_i$
	\STATE $\Delta\lambda_i$ := $\lambda_i - \lambda_{i-1}$
\UNTIL {$\Delta \psi_i < \epsilon$ \AND $\Delta \lambda_i < \epsilon$}
\end{algorithmic}

\end{algorithm}

We now focus on how we initiate our optimization process. We compute the initial feasible orientation $(\vect{x_{c}},\vect{y_{c}},\vect{z_{c}})$ as a look-at orientation applied to all targets. We first compute the vector $\vect{y_c}$ as the average of all normalized vectors $\vect{v_T}$; this will balance targets around the center of the screen. We then compute $\vect{x_{c}}$ and $\vect{z_{c}}$ accordingly (\ie $\vect{x_{c}}=\vect{y_{c}}\times\vect{z_w}$ and $\vect{z_{c}}=\vect{x_{c}}\times\vect{y_{c}}$).
We finally compute the 4d vector $(\rho,\gamma,\psi,\lambda)$ as follows:
\begin{align*}
\rho&=0 & \gamma&=0 \\
\psi &= \measuredangle_{\vect{z_w}}\left[\Pi_{\vect{z_w}}(\vect{y_w}), \Pi_{\vect{z_w}}(\vect{y_c})\right] &
\lambda &= \measuredangle_{\vect{x_c}}\left[\Pi_{\vect{z_w}}(\vect{y_c}), \vect{y_c}\right]
\end{align*}
where the frame $(\vect{x_w},\vect{y_w},\vect{z_w})$ represents the default look-at orientation of the drone (\ie with $\rho=0$, $\gamma=0$, $\psi=0$ and $\lambda=0$).

%% file: interactive-control.tex

\begin{figure*}[t!]
	\centering
	\begin{tabular}{c c c}
		\includegraphics[width=0.24\linewidth]{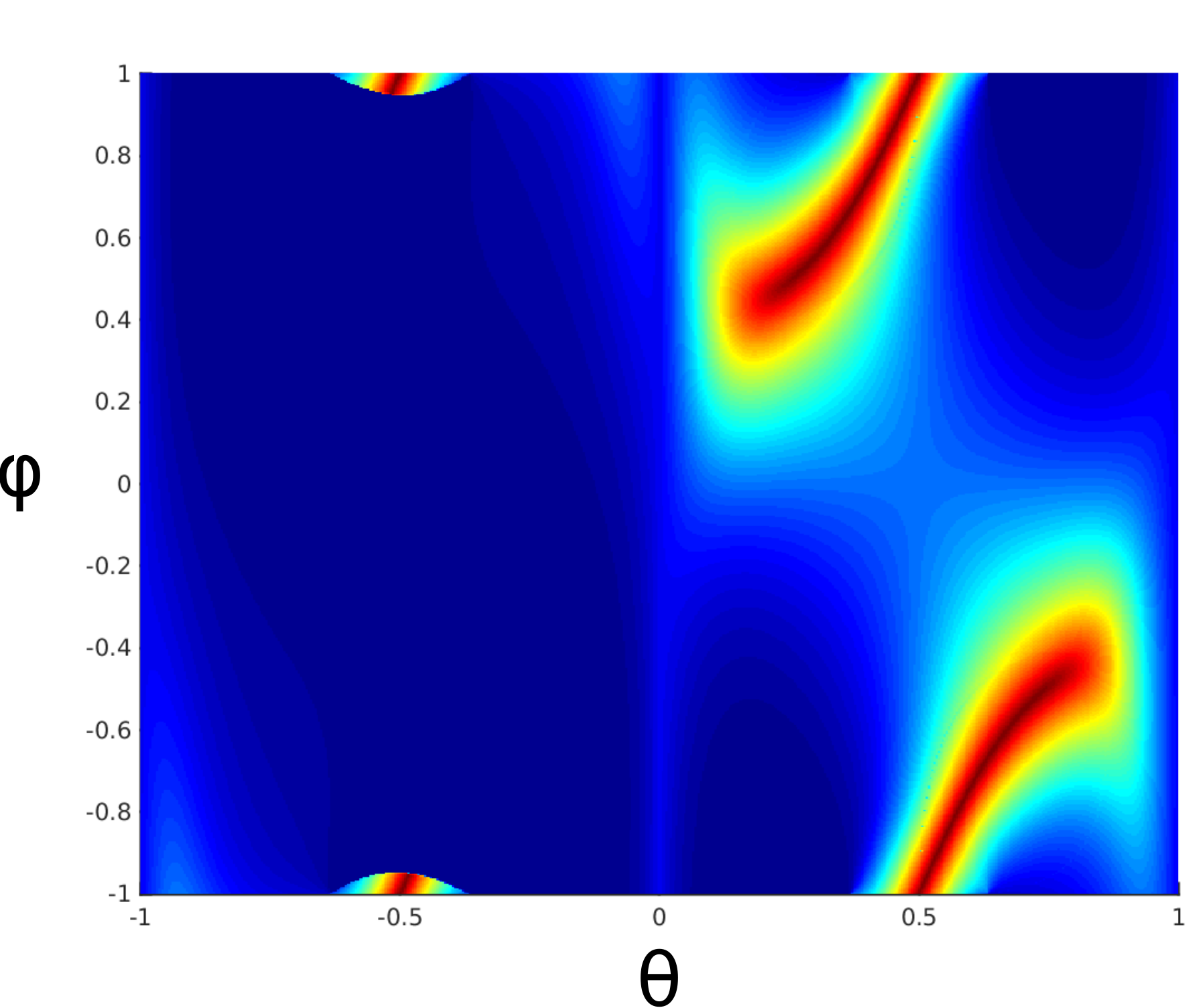} &
		\includegraphics[width=0.24\linewidth]{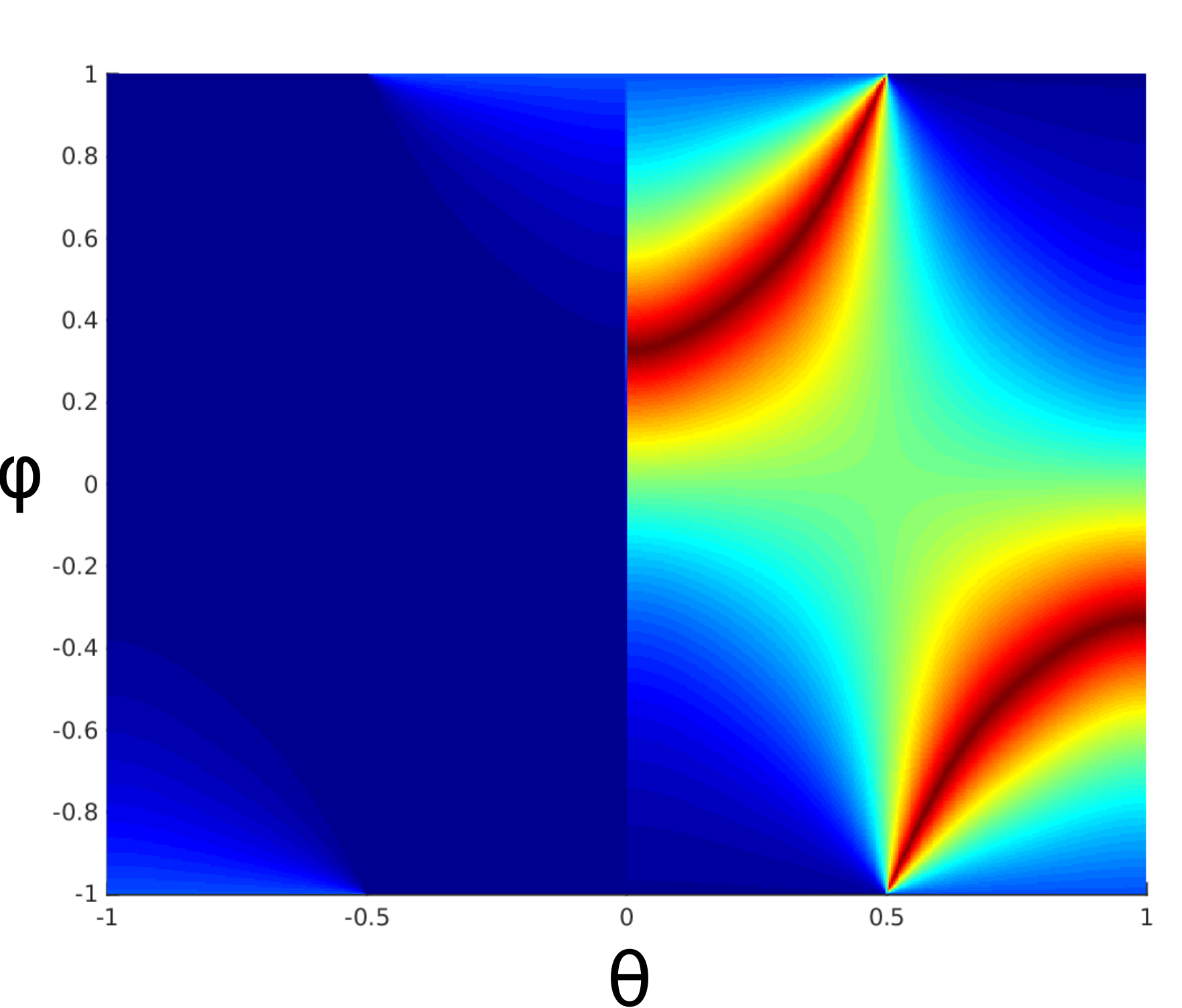} &
		\includegraphics[width=0.24\linewidth]{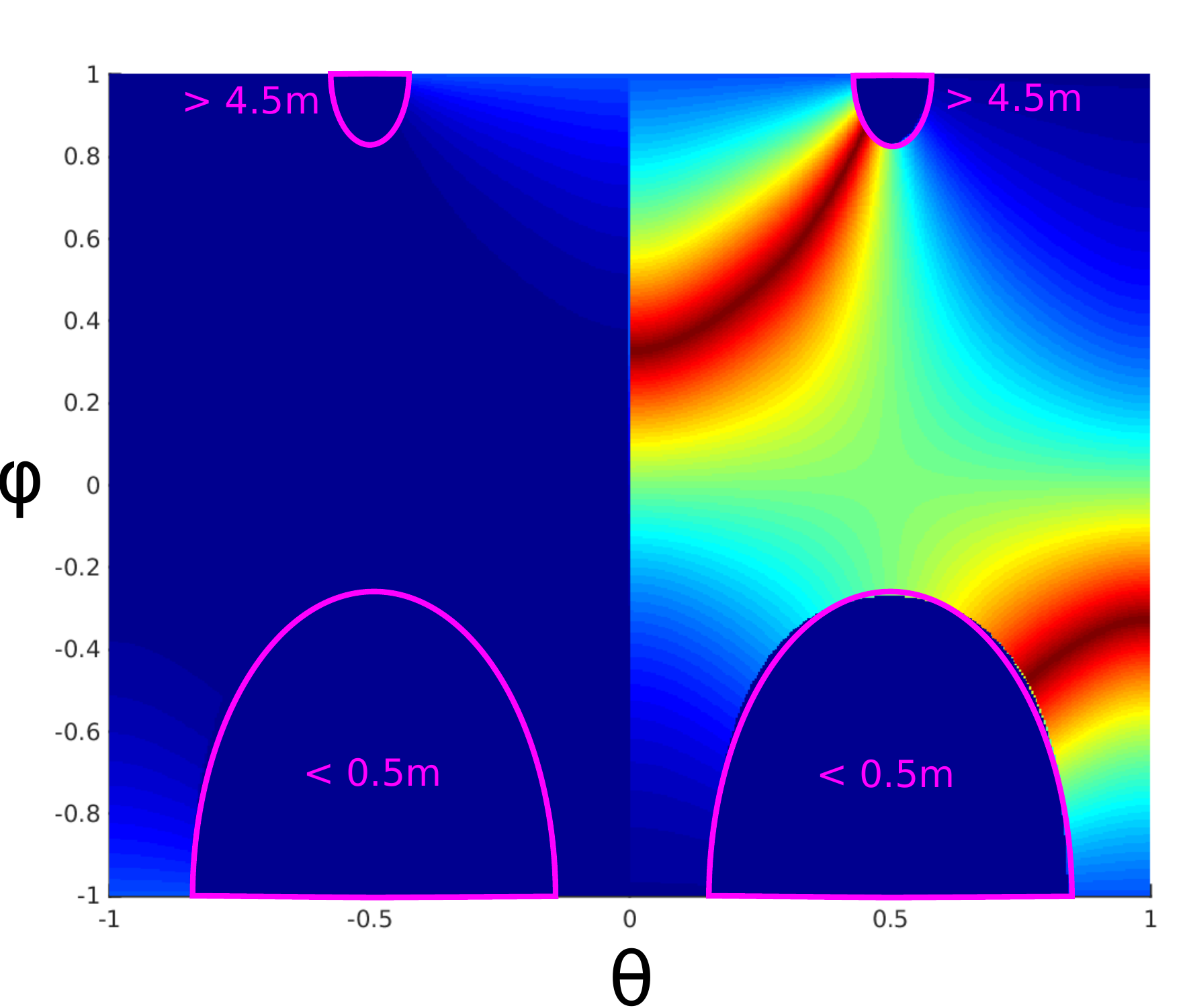} \\
		(a) Lino \& Christie 2015 &	(b) Our satisfaction function &	(c) Floor/ceiling constraints \\
	\end{tabular}
	\caption{Our through-the-lens target repositioning technique needs to rely on a minimization of the drone's roll angle performed in the Drone Toric Surface ($\varphi,\theta$), while satisfying the users' repositioning inputs. Red regions represent the drone positions with a roll value closer to 0, blue regions furthest to zero. As displayed in image (b), our Drone Toric Surface finds solutions with a roll value close to zero even when placed behind target A or B (\ie value $\vartheta=0$ or $\vartheta=1$), a possibility not available in Lino \& Christie [2015] as displayed in image (a). Image (c) displays the intersection of the parametric surface with floor and ceiling boundaries.}
	\label{fig:satisfactionToric}
\end{figure*}

To perform through-the-lens interaction with viewpoints, we propose to implement dedicated manipulation operators, which we have adapted to drones' constraints.

From discussions with an expert drone cinematographer, we have extracted two key behaviors that are strongly expected when manipulating a viewpoint around targets:
\begin{itemize} 
\item when starting from a given viewpoint (with features such as screen position, size, and view angle on targets) and when manipulating one feature, the camera is expected to move while maintaining similar values for the other features;
\item any viewpoint manipulation should be reversible (\ie when manipulating back to the initial screen composition the user would expect the camera to come back to its initial location). While obvious, such features are not ensured by the manipulators proposed in~\cite{lino2015intuitive}.
\end{itemize}
We therefore propose to redefine such manipulators to better fit these behaviors as well as incorporate constraints linked to drones' physical limits.

{\bf View angle manipulator}

As the user manipulates the view angle on the targets, we must enforce the targets sizes and their on-screen framing. This can be ensured by moving the drone onto the current Drone Toric surface. Note that in the specific case of moving onto a surface of type \#1 (concave), when the camera gets close to the alignment of the two targets, the size of the closest target changes abruptly (to ensure the framing). Using such a surface would then create a non aesthetic motion behind this target. In this specific case, we thus propose to extend the safety distance in such a way to instead move onto a surface of type \#3 (\ie the closest convex \emph{DTS}). With this modification, the framing is slightly altered but this avoids unaesthetic motions and preserves the $C^1$ continuity of the camera motion.

\begin{figure*}[t!]
	\centering
	\begin{tabular}{c c c c}
		\includegraphics[width=0.23\linewidth]{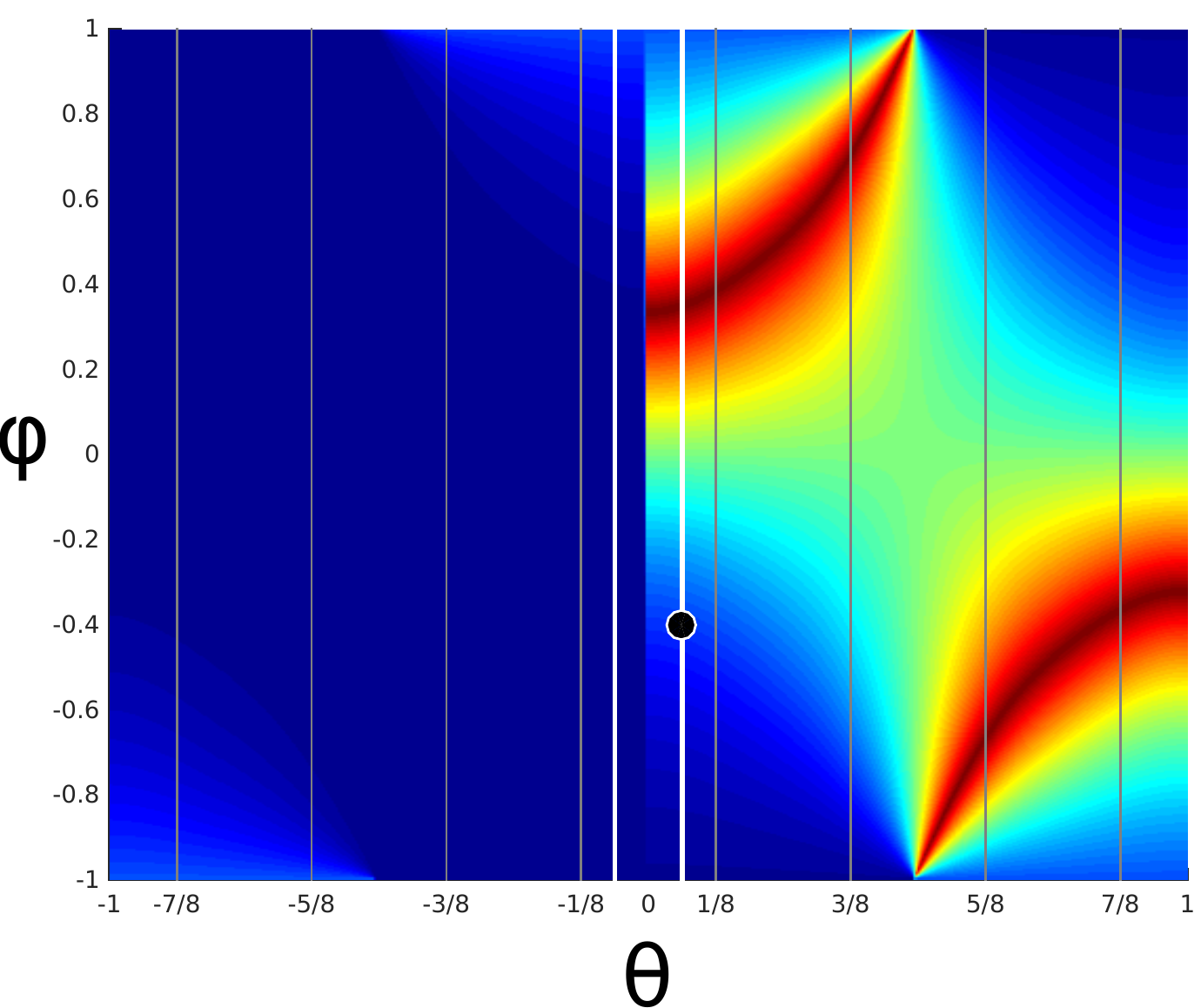} &
		\includegraphics[width=0.23\linewidth]{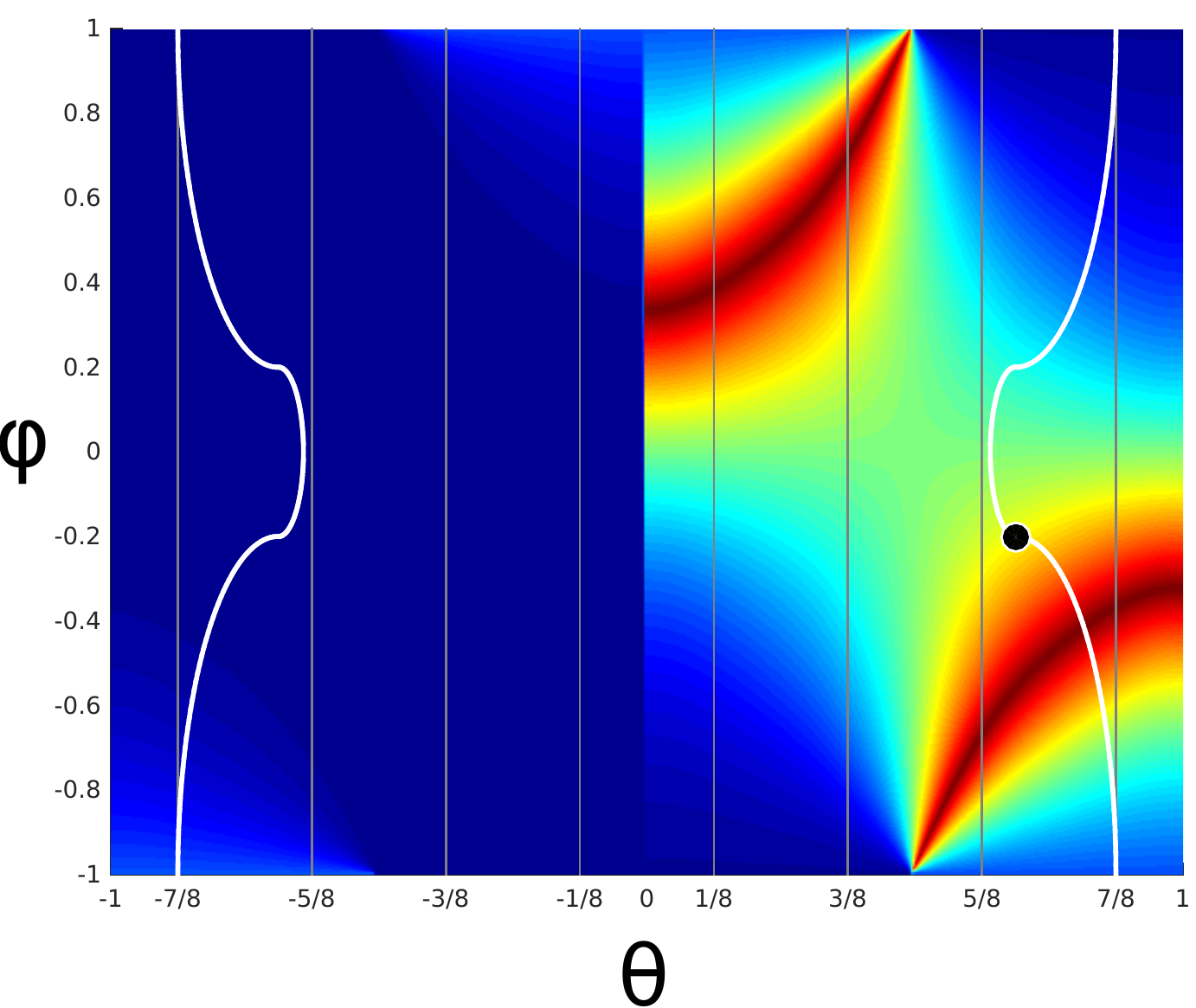} &
		\includegraphics[width=0.23\linewidth]{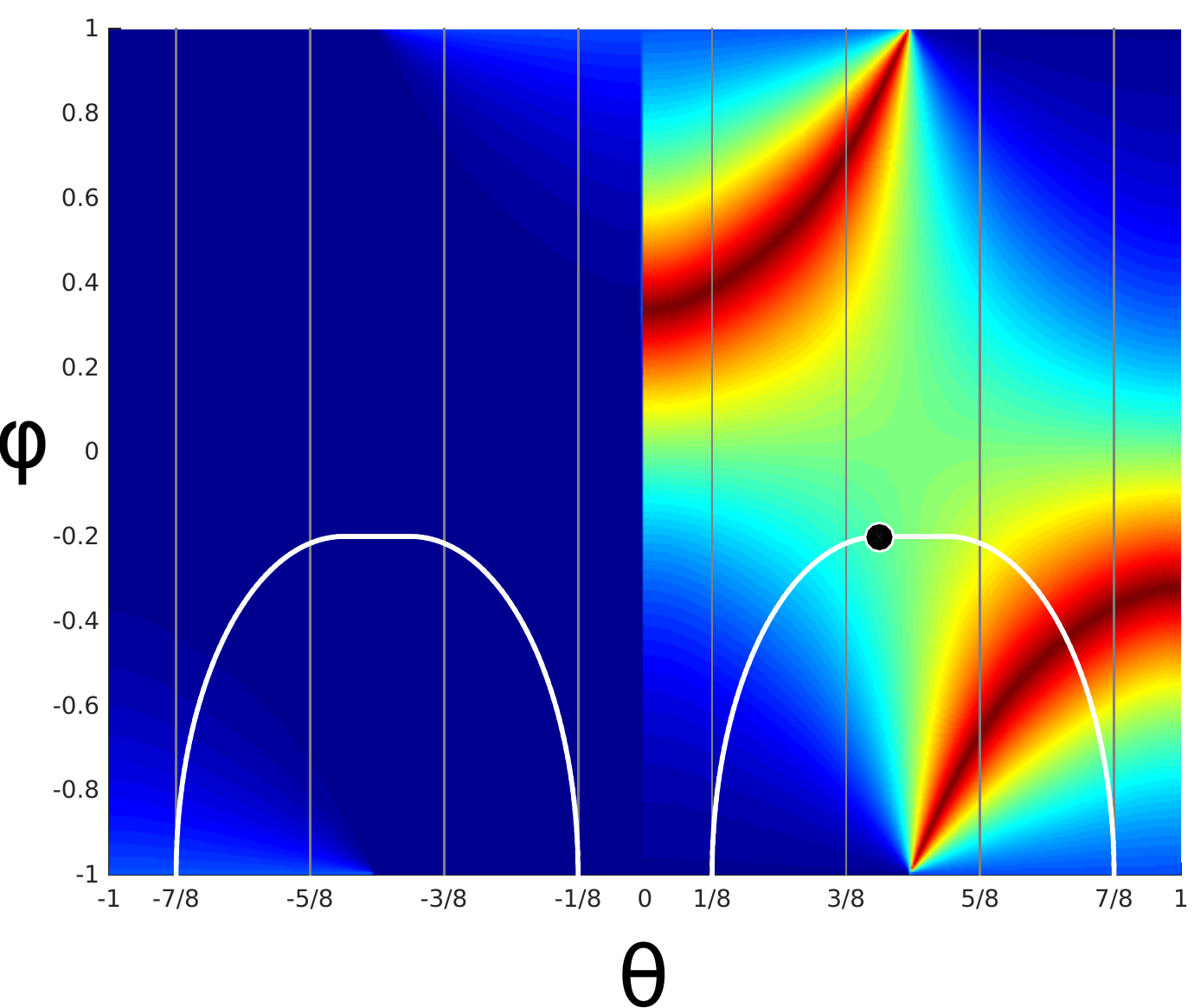} &
		\includegraphics[width=0.23\linewidth]{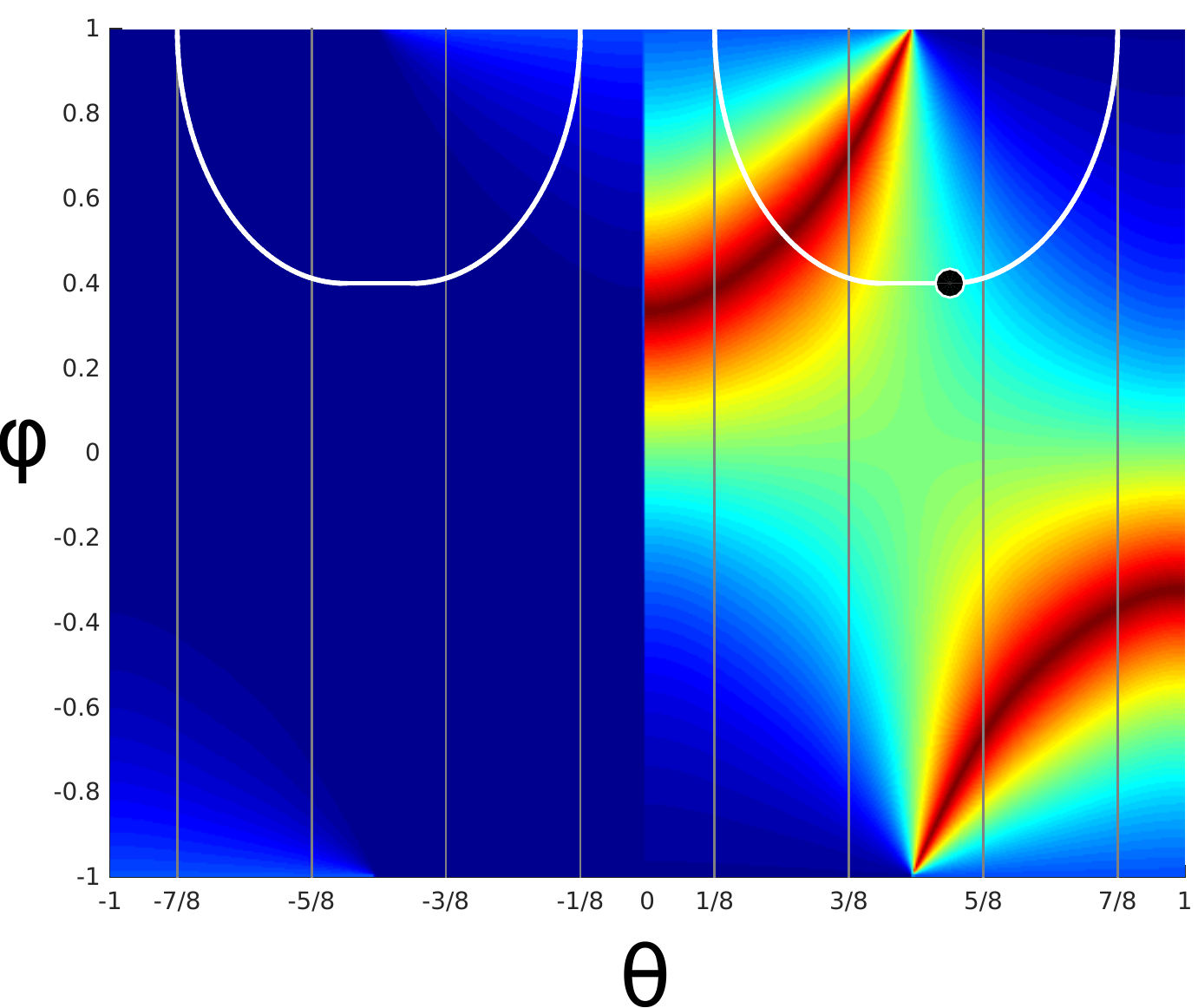} \\
		(a) External B &
		(b) External-Apex A &
		(c) Apex (from below) &
		(d) Apex (from above) \\
		&&& crossing 180$\degree$ line
	\end{tabular}
	\caption{On user manipulation, our optimization process searches from the initial viewpoint (black dot) a drone configuration in the Drone Toric Space with minimum roll. The search is performed in a way to maintain the camera in its original region: external view, apex view, or external-apex view (delimited with gray lines). It tries to avoid conflicts with the floor and ceiling constraints, while enforcing a continuous and soft camera motion and enabling to smoothly cross the 180$\degree$ line (points $(\varphi,\theta)$ and $(\varphi,-\theta)$ coincide in the Cartesian space). To ensure the new viewpoints are coherent with the initial viewing angle, the curve is also restricted to an appropriate interval of values on parameter $\theta$.}
	\label{fig:search-curves}
\end{figure*}

{\bf Position manipulator}

As the user manipulates the framing of one target, we must enforce the framing of the other target, as well as keep the view angle and sizes coherent with the initial viewpoint semantics. Note that, in the case of a single target, (whatever the camera position) the re-framing task is a matter of computing the right camera orientation through the formulas provided in section \ref{sec:feasible-orientation} (\ie the target can always be re-positioned on the screen, without changing the camera's position).

We thereafter only focus on manipulating the framing of two targets, and improving over \cite{lino2015intuitive}.
Note that, in their technique, the computation of a new camera configuration is only accounting for the two targets' framing. This is done through a local search of a viewpoint, that perfectly matches both targets on-screen locations, onto a purely Toric surface.
Therefore, their satisfaction function is not fully suitable in the case of a \emph{DTS}. Due to the generalization, the regions behind targets will not contain perfect solution viewpoints (\ie perfectly matching both targets on-screen locations), as illustrated in figure \ref{fig:satisfactionToric}(a). This could thus prevent from manipulating the camera when it is in such a region (\ie a camera jump would occur). Further, their local search may lead the camera into a local minimum, which the camera may not be able to exit. Their method is thus non-reversible, and does not provide a means to keep the view angle coherent with the initial viewpoint. Finally, their position manipulator does not account for constraints such as keeping the drone in-between a floor and a ceiling height.

To overcome these problems, we here propose several improvements. Firstly, we compute the camera orientation so that, from a given position onto an \emph{DTS}, it best satisfies the desired targets screen locations by using the original viewpoint computation in \cite{lino2012efficient}; in this formulation the roll angle of the camera is left free, conversely to \cite{lino2015intuitive}. We then search, onto the current \emph{DTS}, for a viewpoint with a feasible orientation (\ie whose roll angle is null). With this method, though the framing may not always be perfectly satisfied, we ensure that a feasible drone configuration can always be found, even in regions behind targets (see figure \ref{fig:satisfactionToric}(b)).

Secondly, we account for floor and ceiling constraints. Note that such constraints are difficult to model onto a \emph{DTS} along time. Thus, we do not handle them in a straightforward way, but instead provide a first avoidance mechanism. To do so, we here make two strong assumptions: (i) targets are at the same height in the scene and (ii) in such a situation, a quick analysis (illustrated in figure \ref{fig:satisfactionToric}(c) shows that the set of positions to avoid form perfectly centered ellipses on the top and bottom of the parameter space of a \emph{DTS}. These hard constraints are then enforced more strictly in a further step (see section \ref{sec:collisions}).
Lastly, we must keep the drone in a camera region coherent with the semantics of the initial viewpoint (\ie when starting the manipulation). 

By building upon the above assumptions and constraints, we propose a novel position manipulator which (i) enforces the new target's on-screen positions, (ii) keeps the new viewpoint coherent in terms of the initial view angle, (iii) ensures the manipulation is reversible, and (iv) avoids violating floor or ceiling constraints. We rely on the subdivision into characteristic camera regions to design five control spaces (\emph{external A}, \emph{external-apex A}, \emph{apex}, \emph{external-apex B}, \emph{external B}) for which we ensure a smooth transition between their behavior. Each control space relies on the search of a new viewpoint onto a search curve, bounded to the corresponding camera region (see figure~\ref{fig:camera-regions}) and made of different segments which the design is explained below.
Starting from an initial viewpoint at position $(\varphi_0,\theta_0)$, we then use a dichotomous search (\emph{Golden Section Search}) performed on each segment of this curve separately. This ensures that we can deterministically find a solution, onto the current \emph{DTS}, from any initial viewpoint and manipulation.

For an external view, the search curve is made of two segments (one on each side of the 180$\degree$ line), designed as the lines $\theta=|\theta_0|$ and $\theta=-|\theta_0|$ (which preserve both the sizes and the view angle on targets). Note that, for  external views, floor and ceiling constraints can be considered as negligible.
For an apex or an external-apex region, the search curve is made of a series of ellipse segments that ensure the following features: (i) the search curve contains the initial viewpoint point $(\varphi_0,\theta_0)$, (ii) it is made of a symmetry with regards to the 180$\degree$ line, (iii) it intersects as less as possible the floor and ceiling constraints and (iv) to obtain a $C^1$ curve, adjacent segments are joined at points where their tangents are equal. 
For the apex view, the control space is designed to also overlap the two external-apex regions; this ensures a better collision avoidance while it preserves a similar viewpoint semantics. We also subdivide this control space into two sub-spaces, handled in a symmetric way: apex views from above or from below targets. For external-apex views the control space has a behavior that smoothly transitions between the behavior of the external and apex control spaces.

{\bf Dolly manipulator}
The user can also manipulate the size of a target. The dolly-in / dolly-out manipulator provides a camera motion that makes a selected target bigger or smaller, while it preserves its view angle. This manipulator is implemented as a motion (forward or backward) in the direction of the target.

{\bf World-space manipulators}
To enable more subtle interactions on the viewpoint, we also offer a set of existing world-space manipulators to move the drone forward, left or up, as well as to pan or tilt the drone. These manipulators are common in most 3D modelers; they can be viewed as moving the camera along a rail, with a grue, or rotating it with a pedestal.

\subsection{Collision Avoidance}
\label{sec:collisions}

As we mentioned earlier, a hard constraint is that, during the viewpoint manipulation, we must output a valid drone configuration, \ie a feasible drone configuration that moreover is not colliding nor with a static (object) or dynamic (target) obstacle and is located within the physical bounds of the scene.

\begin{figure}[t!]
	\centering
	\includegraphics[width=\linewidth]{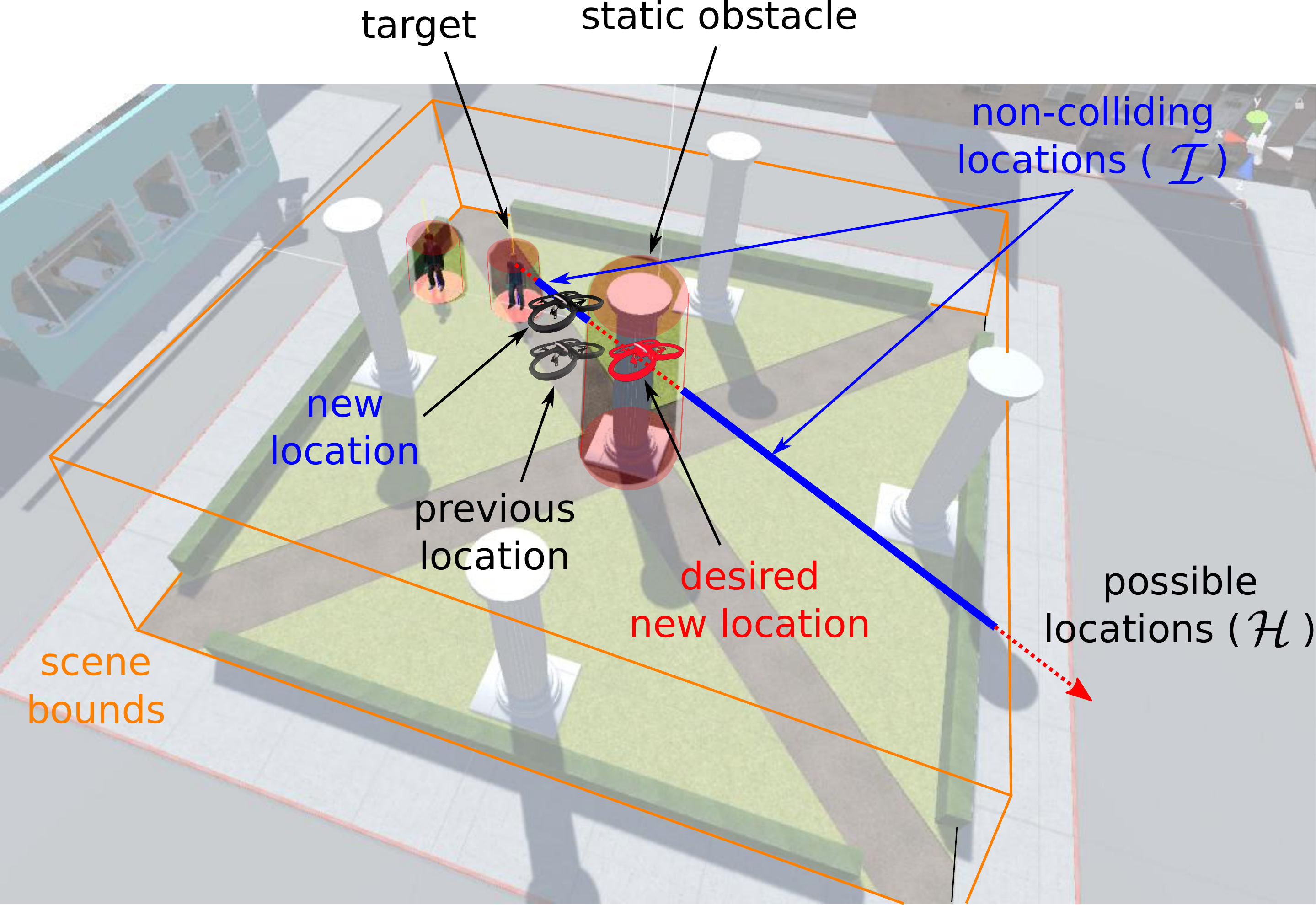}
	\caption{While interactively manipulating the on-screen properties of a target, we strictly avoid collisions by pushing or pulling the drone to a non-colliding location while taking the previous drone location into account.}
	\label{fig:collision-avoidance}
\end{figure}

At this step, we assume a desired new viewpoint has been computed to fit a given screen manipulation (as explained in section~\ref{sec:interactive-control}). If a collision is detected, a natural solution is then to push or pull the drone to avoid this collision. We propose to apply this push/pull mechanism along a half-line $\mathcal{H}$ passing through the desired new viewpoint and whose origin is the target linked to the current screen manipulation (this mechanism is illustrated in figure~\ref{fig:collision-avoidance}). We intersect $\mathcal{H}$ with all (static or dynamic) obstacles and with the bounds of the scene. This provides us with a set of segments $\I$ onto which the drone could be placed along $\mathcal{H}$. Finally, to avoid large variations in the drone location (\eg jumping around an obstacle), we select the new drone location so that (i) it belongs to one segment of $\I$ and (ii) it is as close as possible to the previous drone location.

When no potential location can been found (\ie $\I=\emptyset$), the default behavior is to not move the drone. This ensures safety not only with regards to obstacles but also targets; in particular moving targets then have the possibility to safely avoid the drone.

%% file: planning-paths.tex

The key challenge in filming dynamic targets with drones is to compute cinematographic and collision-free paths. 
To this end, we first propose a method to compute the shortest collision-free path between two drone configurations by introducing a novel composite distance metric expressed partly in the Drone Toric Space (to ensure cinematographic properties over targets) and partly in the Cartesian space (to avoid variations on the altitude of the drone).
We then propose an algorithm to plan a collision-free path from a user sketch, which draws inspiration from~\cite{Gebhardt:2016:AOP:2858036.2858353} but handles dynamic obstacles.
Finally we present a novel path-smoothing technique which, when applied on the computed paths, outputs $C^4$-continuous drone trajectories. 

\subsection{Planning cinematographic collision-free paths}
\label{sec:planning-from-compositions}

The planning of collision-free paths first relies on the choice and computation of a roadmap which samples the free space of a 3D envrionment. Here we rely on a visibility-aware roadmap construction~\cite{oskam2009visibility} in which a prior sphere-sampling stage of the static free space is performed, and the visibility between every pair of spheres is precomputed using raytracing. Adjacent spheres are then connected with arcs in order to construct a roadmap (which we refer to as $\mathcal{R}$). The roadmap is dynamically updated at every frame with moving obstacles (nodes that intersect the obstacles are tagged non traversable).




Generating a qualitative path in this roadmap (in terms of cinematographic properties) rarely means computing the shortest one in the Cartesian space. We propose a novel metric to express shortest paths in the space of visual properties, a shorter path meaning that there are less variations over these properties. Interestingly, our Drone Toric Space already represents visual properties such as distance to targets, angle and composition. Also, to minimize variations in the drone altitude along the path, we extend the space with dimension $z$ (a small variation of $\varphi$ onto a \emph{DTS} may lead to a strong change in altitude). Hence, we cast our path planning problem as a search for the shortest path in the roadmap by expressing nodes in a 4d space $(\alpha,\varphi,\theta,z)$ (which we refer to as the $\tau$-space) and by relying on Euclidian distances between nodes. The cost of an arc between two nodes in the roadmap is then computed by weighting the distance $\tau$-space with visibility of targets. The computation of the path is performed by considering the targets as static. If the targets move, or if obstacles move, the path is recomputed to account for the changes.

Given two drone configurations $\vect{q_s}$ (starting viewpoint) and $\vect{q_e}$ (ending viewpoint) and given targets, the process consists in first adding $\vect{q_s}$ and $\vect{q_e}$ as nodes in the roadmap, and then expressing these configurations in $\tau$-space. We further account for the drone's current acceleration and speed in order to temporarily tag all nodes that would not be reachable by the drone (considering its maximum speed and acceleration) as non traversable. We then rely on a classical A* process to search the roadmap. The distance metric in this search is computed as follows:


{\bf Evaluating the variation in screen properties}. This variation is expressed as the distance between nodes $n_i$ and $n_j$ in the Safe Toric Space. In a practical way, we normalize the lengths and compute the squared distance as: 
\begin{equation*}
D^2_{s}(n_i,n_j) =
\left(\frac{\alpha_i - \alpha_j}{2\pi}\right)^2
+ \left(\frac{\varphi_i - \varphi_j}{2\pi}\right)^2
+ \left(\frac{\theta_i - \theta_j}{2\pi}\right)^2
\end{equation*}

{\bf Evaluating the variation in height} Height difference computed and normal difference in the drone height between both nodes. In a way similar to screen properties, we homogenize this change and compute the squared distance in the world space as
\begin{equation*}
D^2_{h}(n_i,n_j) =
\left(\dfrac{\left| z_i - z_j \right|}{\left| z_h-z_m \right|}\right)^2
\end{equation*} where $z_h$ and $z_m$ represent the highest (resp. lowest) possible positions of the drone. The length of an arc in the $\tau$-space is then expressed as:
\begin{equation*}
D_{\tau}(n_i,n_j) = \sqrt{D^2_{s}(n_i,n_j) + D^2_{h}(n_i,n_j)}
\end{equation*}

{\bf Visibility.} In a way similar to Oskam \etal \shortcite{oskam2009visibility}, we evaluate how much targets are occluded along an arc by using the visibility information encoded in the roadmap. This cost $O(i,j)$ is normalized (0 is fully visible and 1 fully occluded).

The length $L$ of an arc is then defined as:
\begin{equation*}
L(n_i,n_j) = \left[1 + w_o . O(n_i,n_j)\right] . D_{\tau}(n_i,n_j)
\end{equation*}
where $w_o$ defines the weight ($w_o \in [0,1]$) associated to visibility.

\subsection{Planning collision-free sketched paths}
\label{sec:planning-from-sketch}

\begin{figure*}[t!]
	\centering
	\begin{tabular}{c c c}
		\includegraphics[width=0.20\linewidth]{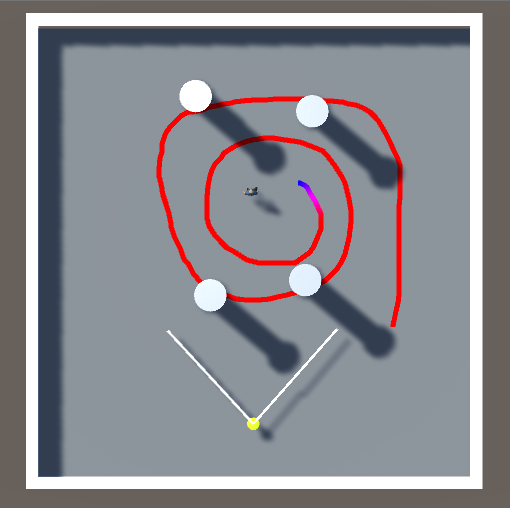} &
		\includegraphics[width=0.20\linewidth]{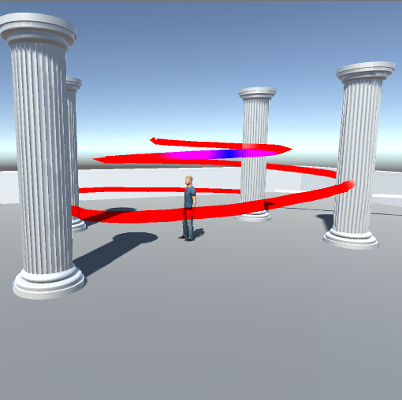} &
		\includegraphics[width=0.20\linewidth]{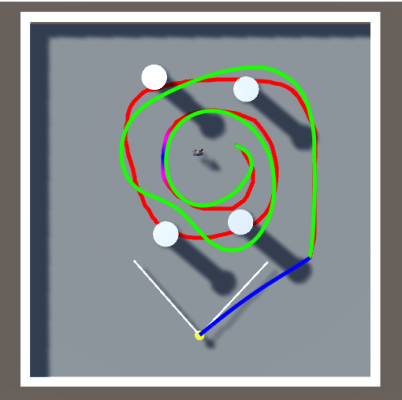} \\
		(a) Drawing the path &
		(b) Adjusting the height and previewing &
		(c) Collision-free path \\
	\end{tabular}
	\caption{Our sketch-based path design: (a) the user sketches a trajectory (in red) without accounting for obstacles, (b) the trajectory's height can be manually adjusted along the path, (c) a collision-free and feasible drone path is computed (in green). A path is also computed from the current drone's position to the beginning of the computed path (blue).}
	\label{fig:drawing-trajectory}
\end{figure*}

We now consider the problem of planning a collision-free path inside our roadmap from a manually sketched trajectory. Gebhardt etal.~\cite{Gebhardt:2016:AOP:2858036.2858353} proposed a similar sketching interaction mode but restricted to static scenes. Our approach tracks dynamic targets when moving along the trajectory and avoids dynamic obstacles. Furthermore, we provide the user with a fine control of the speed at which the drone moves along the path, while caping the speed so that the path remains feasible.



As illustrated in Figure~\ref{fig:drawing-trajectory}, the user sketches a trajectory, refines the height along the trajectory and our system computes a collision-free the trajectory that the user can then execute with a drone. The sketched trajectory (which we refer to as $\mathcal{S}$) is composed of a list of 3D positions. Our goal is then to plan the closest path to $\mathcal{S}$ through the roadmap $\mathcal{R}$. Running an A* algorithm in the roadmap $\mathcal{R}$ would fail due to loops (the user may specify self intersecting trajectories). We proposed a straighforward adaptation through the definition of an oriented graph structure $\mathcal{G}$  which is computed from $\mathcal{R}$ by extending nodes with a value $m$. Each node $v$ (which we refer to as a virtual node) in $\mathcal{G}$ is a pair $(n,m)$  where $n$ stands for the corresponding node in  $\mathcal{R}$ and $m$ stands for an index representing a position in the sketched trajectory $\mathcal{S}$. A virtual node $v_1(n_1,m_1)$ is considered a predecessor of another virtual node $v_2(n_2,m_2)$ if and only if (1) $n_1$ and $n_2$ are directly connected in $\mathcal{R}$ and (2) $m_2>m_1$. 

A traditional A* can then be applied on the structure $\mathcal{G}$: a list of potential nodes (referred to as $\mathcal{P}$) is used, and is updated by adding the neighbors of the current best node $v_b(n_b,m_b)$ (\ie the one having the best potential, which is subsequently removed from $\mathcal{P}$). The potential of a node is evaluated through a cost function and a heuristic function. The process stops as soon as the final node is found or when $\mathcal{P}$ becomes empty. The graph $\mathcal{G}$ in constructed in a lazy way (\ie virtual nodes are only created when evaluated). In practice, we only evaluate successor nodes $v_s$ of $v_b$ for which the value $m_s$ is comprised in the interval $\llbracket m_b+1, m_b+W \rrbracket$ (where $W$ is our search window along trajectory $\mathcal{I}$). One should note that constraining the search into a predefined window ensures that the algorithm detects intentional loops.
We compute the cost of such a successor $v_s$ as
\begin{equation*}
c(v_s) = c(v_b) + (m_s - m_b).\norm{\mathcal{S}[m_s] - n_s}
\end{equation*}
and the heuristic cost of $v_s$ by using an average distance error: $h_c(v_s) = c(v_s)/m_s$





\subsection{Generating $C^4$ continuous paths}


Once we have computed a raw path through our roadmap, we then smooth it so that it is $C^4$-continuous. This is a key criteria for being feasible by a drone as pointed by~\cite{mellinger2011minimum}. Comparatively, Oskam \etal \shortcite{oskam2009visibility} only compute a $C^2$-continuous spline by Hermite interpolation of key-points. Instead, we ensure a $C^4$-continuity by relying on a piecewise polynomial representation of degree 5. These polynomials $\mathcal{P}_i$ are expressed as: 
\begin{equation}
\label{eqn:polynomial}
\mathcal{P}_i(t) = a_i + b_i . t + c_i . t^2 + d_i . t^3 + e_i . t^4 + f_i . t^5
\end{equation}
where $t\in[0,1]$.

The raw path is composed of $N$ key-points $k_i$, each representing a node traversed by this path. Our smoothed path is then composed of $N-1$ polynomials, each satisfying the following constraints:
\begin{align}
\label{eqn:points}
\begin{split}
\mathcal{P}_i(0) ~=~ & k_i,~\forall i \in [1\isep N-1] \\
\mathcal{P}_i(1) ~=~ & k_{i+1},~ \forall i \in [1\isep N-1]
\end{split}
\end{align}
and
\begin{align}
\label{eqn:continuity}
\begin{split}
\mathcal{P}_i^{'}(1) ~=~ & \mathcal{P}_{i+1}^{'}(0),~ \forall i \in [1\isep N-2]  \\
\mathcal{P}_i^{''}(1) ~=~ & \mathcal{P}_{i+1}^{''}(0),~  \forall i \in [1\isep N-2] \\
\mathcal{P}_i^{(3)}(1) ~=~ & \mathcal{P}_{i+1}^{(3)}(0),~ \forall i \in [1\isep N-2]  \\
\mathcal{P}_i^{(4)}(1) ~=~ & \mathcal{P}_{i+1}^{(4)}(0),~  \forall i \in [1\isep N-2] \\
\end{split}
\end{align}

To solve the problem, we take advantage of a roadmap structure similar to ~\cite{oskam2009visibility}. In our roadmap, a node is a portal between two intersecting spheres. This portal is then a disk that represent all possible keypoints through which a path may connect these two adjacent spheres. We therefore cast our problem as the search of all optimal keypoints $k_i$ onto each of the $N$ portals, while ensuring the $C^4$-continuity of the final path. Equation~\ref{eqn:polynomial} defines $6(N - 1)$ unknown coefficients, while Equations~\ref{eqn:points}~and~\ref{eqn:continuity} provide with only $6N - 10$ equations to be solved when considering all continuity constraints. To fully determine our system, we thus also constrain the first and second derivatives of the first and last polynomials to be null, that is :
\begin{equation*}
\mathcal{P}_1^{'}(0) = 0 \text{, } \mathcal{P}_1^{''}(0) = 0 \text{, } \mathcal{P}_{N-1}^{'}(1) = 0 \text{ and } \mathcal{P}_{N-1}^{''}(1) = 0
\end{equation*}
As a result, the solution of this system provides with a $C^4$-continuous path passing through each of the initial key-points. However, depending on the density of nodes in the roadmap, the obstacles in the scenes and the relative configurations (positions and orientations) of the targets, the computed path is rarely qualitative as is (\ie is often implies sudden changes in position, speed and acceleration).
 
While current approaches focus on minimizing the curvature along a path to improve its quality~\cite{oskam2009visibility}, we propose to minimize the changes in its curvature to generate more homogeneous motions. Another advantage of optimizing the curvature variations is that it reduces the use of the drone's actuators which might also result in battery savings.

Our optimization problem can then be viewed as the computation of all coefficients of the $N-1$ polynomials so as to minimize the absolute curvature variations, \ie:
\begin{equation*}
\min \sum_{i = 1}^{N-1}{\int_{0}^{1} \mathcal{P}_i^{(3)}(t)^2 \, \mathrm{d}t}
\end{equation*}
which, after simplification, can be rewritten as
\begin{equation*}
\min \sum_{i = 1}^{N-1}{3.d_i^2 + 12.d_i.e_i + 20.d_i.f_i+16.e_i^2+60.e_i.f_i+60.f_i^2}
\end{equation*}

This sum must be optimized subject to the same constraints as previously (equations \ref{eqn:continuity}), and the following additional constraints to account for the portal radius:
\begin{equation*}
\begin{split}
\norm{\mathcal{P}_1(0) - k_1} ~=~ & 0\\
\norm{\mathcal{P}_i(0) - k_i} - r_i ~\leq~ & 0,~ \forall i \in [1\isep N-1]\\
\norm{\mathcal{P}_{N-1}(1) - k_{N-1}} ~=~ & 0\\
(\mathcal{P}_i(0)-k_i) \cdot n_i ~=~ & 0,~ \forall i \in [1\isep N-1]\\
\end{split}
\end{equation*}
where $n_i$ and $r_i$ are \resp the normal and radius of the $i$\textsuperscript{th} traversed portal.

This nonlinear constrained optimization problem is solved using an Interior Point solver. The $n$ initial key-points computed by the path planning step serve as an initialization.

\subsection{Following $C^4$ paths}

Once a $C^4$ curve $\mathcal{D}$ (defining the drone trajectory) is computed, the task consists of following this curve as closely as possible. 
This is performed using the control scheme proposed in \cite{Fleureau2016}. Based on a Linear Quadratic Regulator, this architecture ensures that the control input sent to the drone will always be feasible. Given the $C^4$ curve as input the drone is able to precisely follow the trajectory (see section \ref{performance} for detailed error measurments).

To ease the synchronization of the drone motions with the scene evolution (\ie to the timing of targets' motions), we devised a second flight mode. 
Given a computed trajectory we allow the user to control the timing and therefore the evolution of the drone along the path. More precisely, we let the user control the acceleration in order to produce smoother drone motions, \ie with a constant velocity, and avoid jerky motions due to the remote controller.
For such time-free trajectory, the previous $C_4$ constraint becomes deprecated. 
To ensure that the path remains locally feasible --\ie the drone does not deviate from the trajectory -- , both the input speed and acceleration are clamped at a maximum value.
As shown in section \ref{performance}, using the same LQR strategy, the error distance to the trajectory remains sufficiently low to ensure user's safety (inferior to 0.4m).

To compute the future positions along the path based on an input acceleration, we start by performing an arc-length re-parametrization of the path so that every point is defined by:
\begin{equation*}
\mathcal{S}(u) \text{, } u \in [0,\mathcal{L}]
\end{equation*}
where $\mathcal{L}$ is the total curvilinear length of the path.
Then for a given time $t$, and a given drone position $\omega$, we compute the curvilinear index $u_t$ along the trajectory so that $\mathcal{S}(u_t)$ is the closest to the drone ; practically this is performed by minimizing \mbox{$\norm{\mathcal{S}(u_t) - \omega}$} such that $u_t$ is taken in a local neighborhood of $u_{t-1}$ and with \mbox{$u_t \ge u_{t-1}$}. From $u_t$, we can finally compute a new goal position $\mathcal{S}(u_t)$ for the drone along the curve using the input speed and acceleration.

%


At run-time, the roadmap is continuously checked for dynamic obstacles that would be onto the planned path (\ie that all nodes onto the path still belong to the free space). When a future collision is detected, the path is recomputed from the current drone configuration.
%
In addition, we also allow the user to interactively adjust the framing of targets. He is provided with an interface where he can, at any time, choose which target(s) should be framed or directly control the yaw and tilt angles of the drone for a finer control of the on-screen composition. 

This process has two main advantages: (i) the user can adjust the execution of the trajectory to the live evolution of the scene (\ie targets' performance) and (ii) the user is relieved from the hard task of manually controlling the drone position and orientation to avoid obstacles while also maintaining a given framing on targets at the same time. 

%% file: coordinating-drones.tex
Our objective is to coordinate the positions and motions of mutliple drones around dynamic targets by (i) covering distinct cinematographic views of the targets at the same time, and (ii) avoiding conflicts between the drones. The challenges to address here, in a system where targets move freely in the environment, are actually to (i) minimize possible conflicts between drones while ensuring a good cinematographic covering of the scene, (ii) dynamically reassigning locations to drones as conflicts arise. 


To tackle these challenges, we first rely on a master/slave relation hypothesis between drones: at any time, the system has a unique master drone (the drone that is currently shooting the scene), and multiple slave drones not currently shooting but either ready to shoot the scene, or moving towards positions where these would be ready to shoot from. The purpose of the slave drones is to propose at any time alternate and complementary viewpoints on the targets while avoiding visibility conflicts with the master drone (ie each slave should not be in the view frustum of the master's camera). This hypothesis draws its inspiration from editing rooms in TV shows where a director controls which camera is the live feed (the master), knows or controls how the other camera are placed (slaves) and decides when to switch to another camera. As with editing rooms, when switching to another camera, ie switching to a slave drone ready to shoot and without conflicts, the slave drone becomes a master drone and and the master becomes a slave.

Then, to ensure the computation of cinematographic views of targets, we propose to empower the drones with elements of \emph{cinematographic knowledge} expressed as a collection of possible \emph{framings}. A \emph{framing}  is a specification of viewpoint properties expressed in the cinematographic language PSL (Prose Storyboard Language \cite{ronfard2015prose}) relative to one or two targets.
These framings correspond to classical shot angles from film literature (eg shots like apex, over-the-shoulder, medium close-up).

As the targets evolve, the drones move to maintain the framing features. On onset of conflicts, a dynamic reassignment process is performed which minimizes the number and cost of changes to perform using local repair techniques. In the sequel, we present the details of this approach.

\subsection{From framings to framing instances}

To each drone, master or slave, is associated some \emph{cinematographic knowledge}, expressed as a collection of 17 possible \emph{framings}. A \emph{framing} $f$ is a specification of viewpoint properties expressed in the cinematographic language PSL (Prose Storyboard Language \cite{ronfard2015prose}) relative to one or two targets. These framings correspond to classical shot angles in film literature (eg over-the-shoulder shots, medium close-up shots, apex shots). While multiple languages have been proposed for the purpose of controlling a virtual camera \cite{bares2000composition,halper2001engine,ranon2014improving}, only the PSL specification language is strongly tied to cinematography. Designed for both annotating film shots and expressing specifications for computational cinematography, the language only finds partial implementations as \cite{ronfard2015prose,galvane2014narrative}. 

For a framing $f$ with a list $l$ of targets, the positions of which are known, we define a \emph{framing instance operator} which computes a geometric instance of the framing $f$. This instance $I_{f,l}$ is computed and expressed as a volume of possible drone positions in which each position $\tau$ in the Drone Toric Space shares the cinematographic properties of $f$. The idea is founded on the notion of Director Volumes~\cite{lino2010cinematography} which represent a convex volume to which a multiple semantic tags are associated, each tag representing a visual cinematographic property such as visibility, camera angle, shot size, etc. The dynamic computation of these regions was performed using BSPs, a computationally expensive process to be performed in real-time especially when considering visibility with complex scene geometries. In contrast, we express this volume as a convex 3D region defined in the Safe Manifold Surface coordinates. A framing instance $I_{f,l}$ corresponds to a region $\mathcal{\tau} =\langle \boldsymbol{\varphi, \theta, \alpha} \rangle$ where $\boldsymbol{\varphi, \theta, \alpha}$ are intervals of values. 17 distinct regions around one and two targets are defined, each region corresponding to cinematographically distinct framing. In our case, the bounds of each region have been easily designed by hand from the film literature (ie setting $\boldsymbol{\varphi, \theta, \alpha}$).

\subsection{Conflicts between drones}

Three types of conflicts with two natures of conflicts were identified. \emph{Hard conflicts} should be avoided at any time, and \emph{soft conflicts} should be avoided whenever possible, and introduce some flexibility in a problem that otherwise easily becomes over-constrained. 

Collision conflicts are always hard conflicts and enforce a minimal distance between two drones to avoid perturbations due to air thrusts and ground effects. Collision conflicts also occur between the drone and environment constraints (boundaries of the scene or scene geometry). Collision conflicts are handled by performing Euclidean distance computations (in case of collisions between drones), and queries in the roadmap in case of collisions with the static scene geometry.

\begin{figure} [t!]
	\centering
	\includegraphics[width=0.45\linewidth]{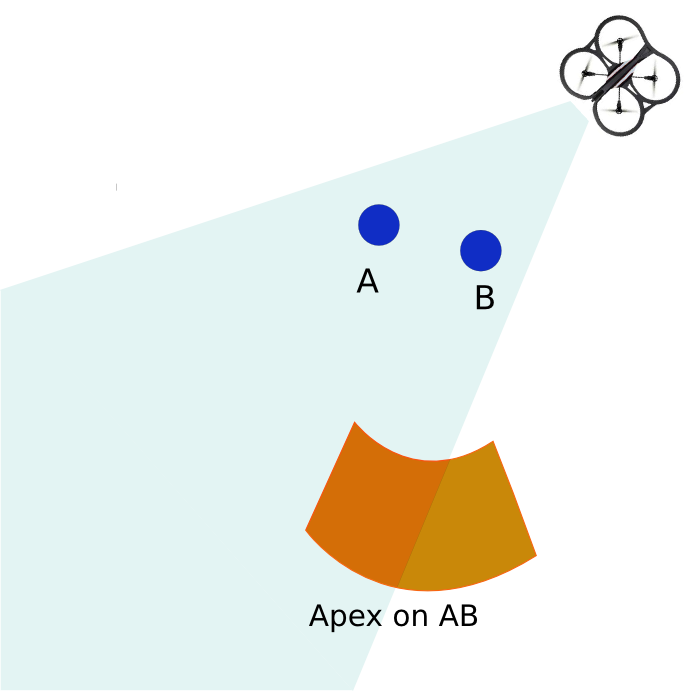}
	\includegraphics[width=0.45\linewidth]{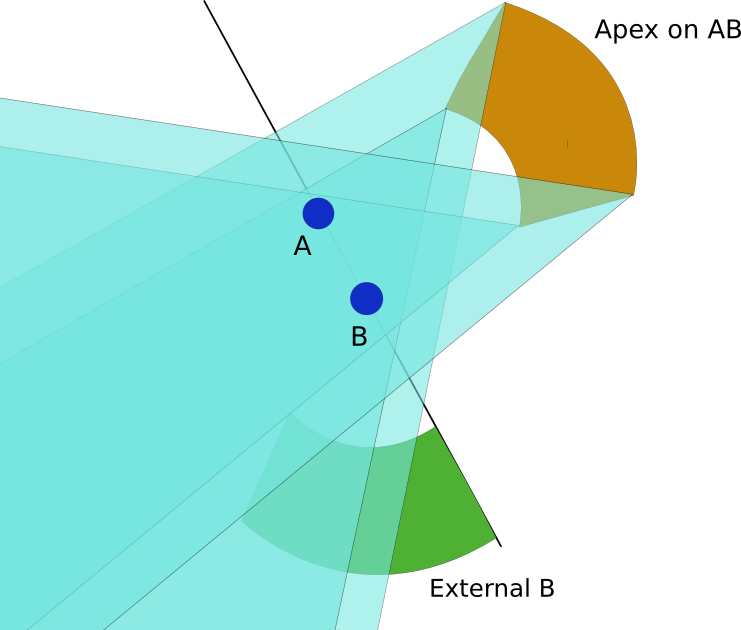}
	\caption{Visibility conflicts are detected using extreme points of the framing instances. The intersection between a frustum and a region is computed using a marching cube dichotomous search in the 3D Drone Toric Space.}
	\label{fig:visibility-conflicts}
\end{figure}

Visibility conflicts (having a drone viewing another drone in it's viewport) is a hard conflict when applied to the master drone (no other drone should be visible in its viewport). But it is a soft conflict between slave drones, or between a slave drone and the master drone (\ie the slave may have the master in it's viewport). Visibility conflicts are detected in a straightforward way when computed between drone configurations using frustum tests. However visibility conflicts also need to be computed between two frame instances (when multiple drones need to select frames instances (\eg during the initial assignment), or between a frame instance and a done position (see Figure~\ref{fig:visibility-conflicts}). Conflict detection is performed in two different stages: (i) when assigning framings to each drone (region to region visibility computation), and (ii) when the scene is evolving (drone to region visibility computation) to check if a region is still valid or to evaluate valid regions. Since the intersection between the framing instance region and the frustum cannot be performed algebraically, a straightforward dichotomous search is performed along the edges of the 3D Drone Toric Space (similar to the \emph{marching cube} technique). For each edge, we approximate the possible intersection with the frustum. A region is then fully visible by another region, partially visible, or non visible at all. 

\subsection{Initial assignment using min-conflict}

Ideally, all conflicts should be avoided between all the drones at any time. However, given the wideness of the view angles we consider (diagonal angle is 92° on the parrot, and 94° on the DJI Phantom 3), the dynamic orchestration of multiple drones around moving targets in a constrained environment while preventing all conflicts quickly becomes in practice an intractable problem. Hence, the motivation behind our master/slave hypothesis  --and beyond the idea of reproducing an TV show editing room-- is also to avoid locked situations where too many simultaneous conflicts limit the possibilities and reduce the practical applicability of the approach and as a side effect, to reduce the overall computational complexity due to pair-wise conflict testing between drones.

The overall approach consists in selecting consistent framings for the master and all the slave drones, so as to avoid visibility conflicts with the master, collision conflicts between all the drones, and minimize angle and visibility conflicts between the slaves. 
This is a straightforward combinatorial assignment problem, easily expressed as a CSP (Constraint Satisfaction Problem). However, we aim at more than just a consistent assignment. First, when there are multiple consistent assignments possible (ie multiple solutions where there are no conflicts), it is preferable to select the one that requires the least energy for the global system (\ie selecting among possible solutions the one for which the total length between current drone positions and target regions are the shortest). Second, when there is no consistent assignment, it is preferable to select an assignment that minimizes the conflicts. To this end, we rely on a min-conflict local search technique~\cite{minton1990minconflict}, which from a first initial assignment,  iteratively selects the drone which has the most conflicts and for this drone selects a candidate framing which minimizes all conflicts. Interestingly the min-conflict strategy can provide a locally best solution at any time.

The process can be formalized as follows. Let's define a drone $d$ to which can be associated one camera specification $f$ among a set of possible specifications $\mathcal{F}$. We provide a function $c(d)$ which computes the number of conflicts the drone $d$ has with other drones, and a second function $f(d_i,s_j)$ which computes the cost for a drone $d_i$ to move to a region corresponding to a specification $s_j$. This cost is is the length of the path in the roadmap from a the drone's position to the center of it's destination region.  Computing $\max{i} c(d_i)$ selects the drone with most conflicts, and the largest cost. The selection of the best framing candidate then relies on searching for the framing $j$ that minimizes $\min{j} c(d)+f(d_i,s_j)$.

\begin{figure} [t!]
	\centering
	\includegraphics[width=0.99\linewidth]{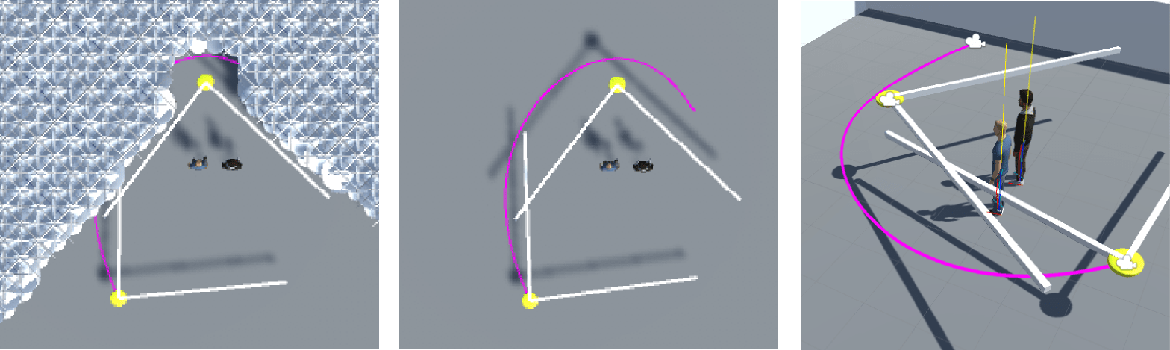}
	\caption{When coordinating multiple drones, the roadmap is dynamically updated by tagging nodes inside the master drone's frustum as non traversable. If a drone is inside the view frustum of the master drone, a path is computed which avoids non traversable nodes.}
	\label{fig:planning-with-frustum}
\end{figure}

Once an initial assignment is performed, we decide a destination position in each region, computed as the center of the region in the Drone Toric Space coordinate system, and then converted into a drone configuration. When regions are partially visible (\ie intersect a frustum), the center of the largest visible volume is computed in the Drone Toric space. We then rely on our drone path planning technique (Section~\ref{sec:planning-paths}) to compute a path to the destination position. To this end, the roadmap is dynamically updated by tagging nodes inside the frustum of the master drone as non traversable, so that slave drones trajectories do not cross the frustum (see Figure~\ref{fig:planning-with-frustum}).

\subsection{Dynamic assignment using local repair}
 
 At any time, as the scene evolves (ie the master drone moves or targets move), the systems maintains the camera framings when possible. When targets move, the framing instances are updated and the drones move towards the center of their updated framing instance. As conflicts appear (eg a framing instance is in conflict), a new combination of framings needs to be assigned to one or multiple drones. To this end, we rely on local repair techniques, a well-known heuristic in dynamic planning  problems~\cite{Miguel2004Dynamic} which minimize the amount of changes in assignments. The process is the following: a list  $\mathcal{L}$  containing the drones in conflict at time $t$ is created. The minconflict is then applied to the drones in $\mathcal{L}$ and gives a subset list $\mathcal{L}'$ of the drones still in conflict. All the drones in conflict with each drone of $\mathcal{L}'$ are then added to the list $\mathcal{L}$ over which min-conflict is re-applied. The process ends when no conflicts are found, or when the minconflict has been applied to all the slave drones in the scene. In the best case, only the salve drones in conflict will be re-assigned a new framing. In the worst case, all slave drones will be reassigned a framing.

%% file: results.tex


To evaluate our system, we used a set of Parrot ARDrone2 equipped with an onboard camera. While these drones do not provide any form of stabilizing gimbal -- which also prevents from controlling the tilt of the camera -- they currently remain the only drone available on the consumer market that are safe enough to be used at very close range.
The experiments were conducted in a 30$\times$20 meter room where a Vicon motion capture system was set up to manage the live tracking of dynamic objects in the scene. The final tracking volume was approximately 15$\times$10$\times$5 meter. Different configurations of this volume were created with various static obstacles placed in the scene. For each configuration, a 3D model of the scene was built and the visibility roadmap was generated as explained in section \ref{sec:planning-paths}. 

Our tool is designed as a Unity 5 plugin under Linux, linked to the control system of each drone through a Wi-Fi connection. It also provides a \emph{simulation mode} allowing to simulate the drone behavior, based on its physical model. With this mode a user can train on the tool, without having to fly the drone, and has the possibility to interactively move targets and obstacles in the 3d scene.

\subsection{User evaluation}
\label{userEval}

We here evaluate the relevance of our tool to create drone shots, in a known environment with static and dynamic obstacles.  
We have conducted a user study, in which we have compared both our image-space manipulation tool (referred to as the \emph{framing tool}, described in section \ref{sec:planning-from-compositions}) and our \emph{sketching tool} (described in section \ref{sec:planning-from-sketch}) with a traditional drone control device -- in our study, we used a controller with two analog sticks. 
We recruited twelve novice users with no prior knowledge in cinematography nor drone piloting. 
After a short demonstration of the three modes (\ie framing, sketching and manual piloting), they were given 10 minutes of training on each mode. They were then assigned a series of 3 tasks, to perform both manually (M) and with our tool (T).
Firstly, given a target on-screen composition of two actors (\ie a screenshot), they had to (i)$_M$ manually maneuver the drone and (i)$_T$ use our \emph{framing tool}, to move the drone to a  viewpoint closely matching this target composition. 
Secondly, they had to maintain a given framing over the actors as they moved around in the scene, with (ii)$_M$ the manual mode and (ii)$_T$ our \emph{framing mode}.
Thirdly, to evaluate our \emph{sketching mode}, we asked the participants to perform a series of trajectories (zoom in, traveling, turn around the actors) while maintaining the camera oriented towards the actors. They had to complete it by (iii)$_M$ manually flying the drone with the controller and (iii)$_T$ by using our \emph{sketching tool}.
Finally, each participant had to complete a full questionnaire.
For all three tasks, they had to evaluate -- on a scale from one (worst) to five (best) -- their familiarity with each type of tool, the ease of use, the perceived fluidity of the created camera motions, the precision and their personal satisfaction over the obtained results. In addition, they all provided verbal feedback on each mode.

\begin{figure}
\begin{subfigure}{0.32\linewidth}
\includegraphics[width=\linewidth]{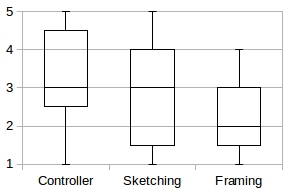}
\caption{Familiarity}
\label{fig:familiarity}
\end{subfigure}
\begin{subfigure}{0.32\linewidth}
\includegraphics[width=\linewidth]{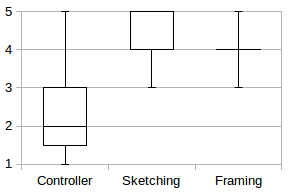}
\caption{Ease of use}
\label{fig:easiness}
\end{subfigure}
\begin{subfigure}{0.32\linewidth}
\includegraphics[width=\linewidth]{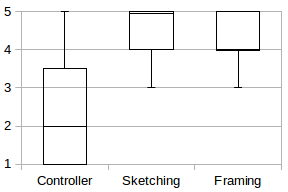}
\caption{Fluidity}
\label{fig:fluidity}
\end{subfigure}

\begin{subfigure}{0.32\linewidth}
\includegraphics[width=\linewidth]{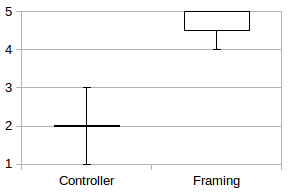}
\caption{Framing}
\label{fig:maintainFraming}
\end{subfigure}
\begin{subfigure}{0.32\linewidth}
\includegraphics[width=\linewidth]{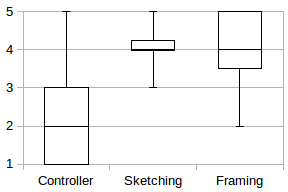}
\caption{Precision}
\label{fig:precision}
\end{subfigure}
\begin{subfigure}{0.32\linewidth}
\includegraphics[width=\linewidth]{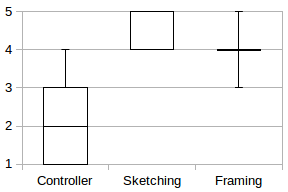}
\caption{Satisfaction}
\label{fig:satisfaction}
\end{subfigure}
\caption{Results of our user evaluations, conducted with 12 participants, to compare manual flight with our sketching and framing tools. On a scale from one to five, each participant evaluated their degree of familiarity with each type of tool (a), their ease of use (b), the fluidity of the drone motion (c), the ease to maintain the framing over one or a group of actors (d), the precision of the tool (e) and their satisfaction over the resulting shot (f).}
\label{fig:results}
\end{figure}

Despite most participants were more familiar with traditional manual controllers (see Figure \ref{fig:familiarity}), they clearly favored our tool on every aspect (see Figures \ref{fig:easiness} to \ref{fig:satisfaction}). In addition to the graded evaluation, many expressed their satisfaction over the results and they found both that our tool easier to use and that its resulting output is more precise (\ie it better fits their intentions). Some referred to our tool as \emph{very promising}, \emph{very interesting} or \emph{fun}.
One drawback of our \emph{framing mode}, highlighted by some participants, is the lack of control on the drone trajectory. Some suggested that the addition of means to specify series of key shots could help better craft trajectories.
In the sketching mode a complaint was that, though the height of the sketched path can be manipulated, it is not possible to make modifications on its shape. Users must re-draw a new path. However, considering this path can be modeled as a spline curve, we could easily include spline manipulators in our \emph{sketching tool}.

From this study, we make several other observations. Not surprisingly, none of the participants were able to produce qualitative footage by manually flying the drone --\ie tasks (i)$_M$, (ii)$_M$ and (iii)$_M$ -- in a simple constrained environment. As we expected, this task would actually require much more training.
Participants were also able to complete tasks (i)$_T$, (ii)$_T$ and (iii)$_T$ in just a few trials. Novices were in general more comfortable with the sketch-based approach, as it allows more freedom than the screen-space manipulation mode. 
Further, all participants provided good feedback. Novice were particularly surprised to see how easily they could produce good-looking footage despite their inexperience in both cinematography and drone piloting. 

\subsection{Expert feedback}

In a separate session, we invited an expert cinematographer (film producer and camera drone pilot) to experiment our tool. After a rapid overview of the tool, he was given 30 minutes to take over the different interaction modes. He was then asked to perform the same tasks as novice users (see section \ref{userEval}). We was finally asked to provide us with his feedback on this experiment. 
Following his advices, we also tested our tool on a variety of other scenarios (the corresponding footage are provided in the accompanying video). These scenarios demonstrate our contributions: onscreen manipulation, cinematographic path planning from screen composition, dynamic obstacle avoidance and drone coordination. These tests were performed in the same environment, comprising multiple obstacles, up to three moving targets, and up to three quadrotor drones.

The expert especially appreciated having to interact with the drone through only a single degree of freedom (its velocity) compared to the 7 degrees of freedom he usually has to interact with when manually flying the drone. He also appreciated not having to handle the avoidance of obstacles, and being able to easily and precisely control the framing. 
He however expressed some frustration due to the unavailability of a simple controller for the tilt of the camera; and he found the creation of an extremely precise trajectory difficult with our sketching tool, when actors are dynamically evolving within the environment. More specifically, his main issue was the precise synchronization of the drone motion with the actors motions during a staged sequence -- in a way similar to a real camera operator, which is currently almost impossible for a single drone pilot.

\begin{figure}
\begin{subfigure}{\linewidth}
\includegraphics[width=\linewidth]{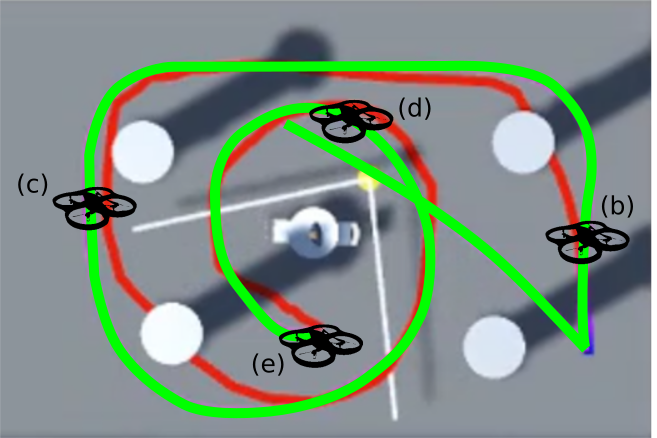}
\caption{Virtual top view of the scene}
\label{fig:topView1}
\end{subfigure}
\begin{subfigure}{0.49\linewidth}
\includegraphics[width=\linewidth]{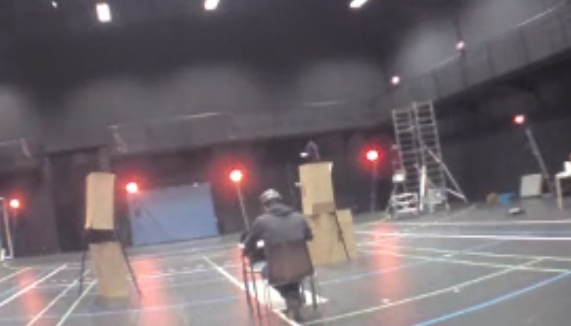}
\caption{}
\label{fig:liveShot1}
\end{subfigure}
\begin{subfigure}{0.49\linewidth}
\includegraphics[width=\linewidth]{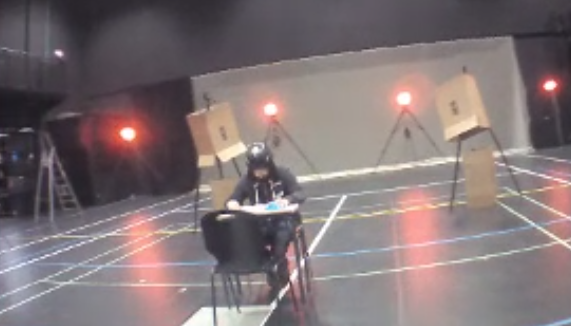}
\caption{}
\label{fig:liveShot2}
\end{subfigure}
\begin{subfigure}{0.49\linewidth}
\includegraphics[width=\linewidth]{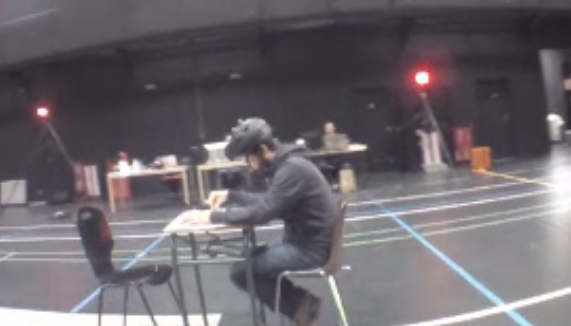}
\caption{}
\label{fig:liveShot3}
\end{subfigure}
\begin{subfigure}{0.49\linewidth}
\includegraphics[width=\linewidth]{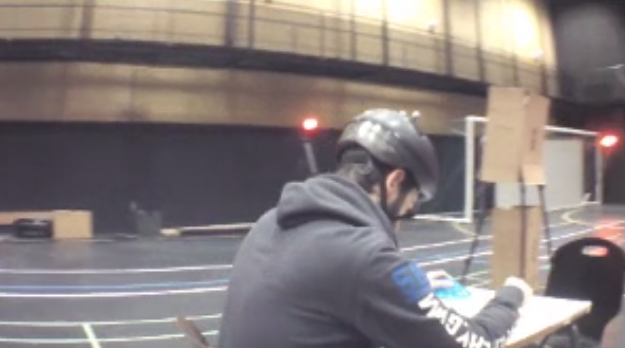}
\caption{}
\label{fig:liveShot4}
\end{subfigure}
\caption{Shot captured with the drone as the expert is flying a trajectory crafted with our sketch based tool. The top view (a) shows the trajectory computed (in green) from the drawn trajectory (in red). (b), (c), (d) and (e) give different viewpoints taken at different times along the trajectory.}
\label{fig:example1}
\end{figure}

\begin{figure}
\begin{subfigure}{0.49\linewidth}
\includegraphics[width=\linewidth]{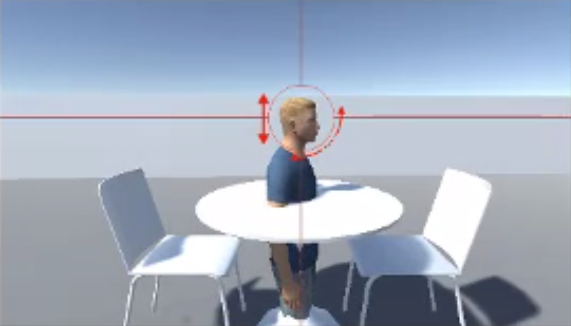}
\label{fig:desiredShot}
\caption{Desired shot composition}
\end{subfigure}
\begin{subfigure}{0.49\linewidth}
\includegraphics[width=\linewidth]{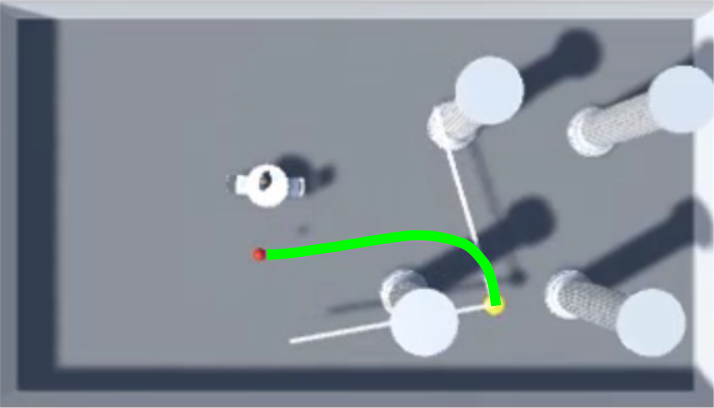}
\label{fig:topView2}
\caption{Virtual top view of the scene}
\end{subfigure}
\begin{subfigure}{0.49\linewidth}
\includegraphics[width=\linewidth]{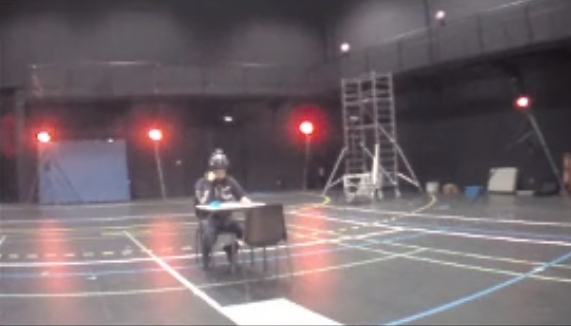}
\label{fig:initFraming}
\caption{Initial framing}
\end{subfigure}
\begin{subfigure}{0.49\linewidth}
\includegraphics[width=\linewidth]{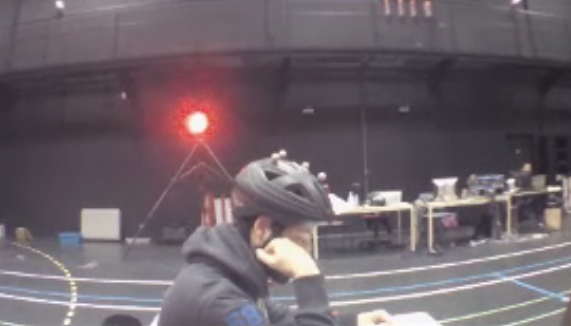}
\label{fig:finalFraming}
\caption{Final framing}
\end{subfigure}
\caption{Shot captured with the drone as the expert is flying a trajectory computed with our framing tool.}
\label{fig:example2}
\end{figure}

Footage from these experiments are included in the companion video and illustrated in Figures \ref{fig:example1} and \ref{fig:example2}.

\subsection{Qualitative evaluation}

We also conducted an objective study, to assess the relevance of our Drone Toric Space compared to a Euclidian space (as used in most previous techniques) in planning cinematographic paths. We extracted and analyzed the evolution of visual properties along a drone trajectory, performed with the different modalities. 
Similar to task (i)$_T$ in section \ref{userEval}, we computed a drone path (from its current position to a desired viewpoint) by using our path planning algorithm (\ie the shortest path in Toric space) and compared the result of the method in \cite{oskam2009visibility} (\ie the shortest path in world space). For the sake of the evaluation, we also asked a user to manually perform the same path.
The experiment was conducted within the simulator. This guaranteed that the initial state of the drone and actors were always the same, thus ensuring that results were not affected by external stimuli.

\begin{figure}
\includegraphics[width=\linewidth]{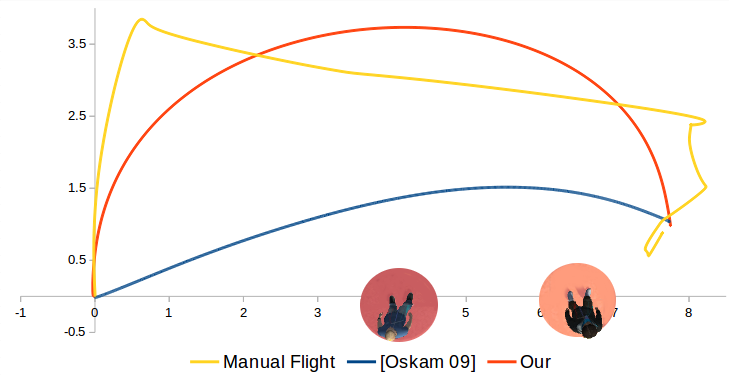}
\caption{Trajectories followed by the drone to reach a specific viewpoint while framing two actors, in manual mode, with \protect\cite{oskam2009visibility} and with our path planning solution.}
\label{fig:trajectories}
\end{figure}

\begin{figure}
\begin{subfigure}{0.32\linewidth}
\includegraphics[width=\linewidth]{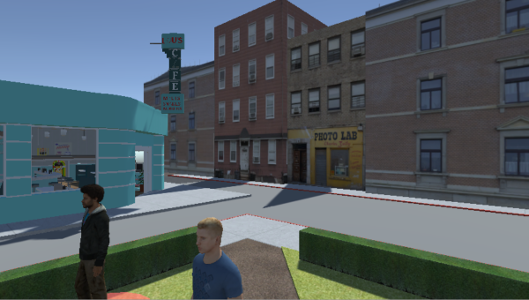}
\caption{}
\label{fig:visibilityShortest}
\end{subfigure}
\begin{subfigure}{0.32\linewidth}
\includegraphics[width=\linewidth]{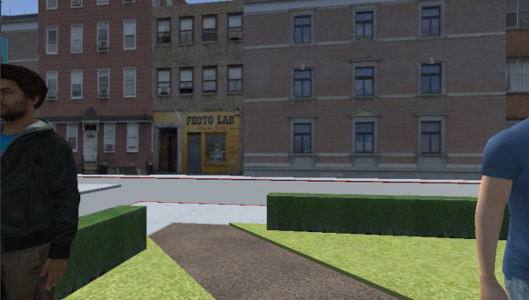}
\caption{}
\label{fig:visibilityManual}
\end{subfigure}
\begin{subfigure}{0.32\linewidth}
\includegraphics[width=\linewidth]{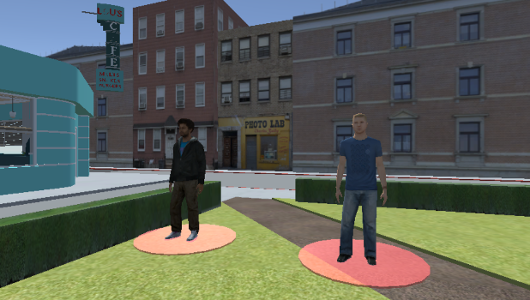}
\caption{}
\label{fig:visibilityCinema}
\end{subfigure}
\caption{Shots captured with the simulated drone as the user tries to reach a given framing in manual mode (a), with the method proposed in \protect\cite{oskam2009visibility} (b) and with our cinematograpic path planning solution (c).}
\label{fig:snapshots}
\end{figure}

Figure \ref{fig:trajectories} shows the performed trajectories in each scenario. As expected, the automated solutions produce smooth trajectories and avoid obstacles (\ie the actors). Conversely, the manual flight resulted in a very noisy trajectory. In terms of visual satisfaction, Figure \ref{fig:snapshots} highlights the main issues encountered with the manual flight and the shortest path solutions. 
When manually flying the drone, the user had difficulties to handle the drone position and the framing of the actors simultaneously. When using \cite{oskam2009visibility}, the drone is not able to maintain the visibility of the actors during the whole flight. As illustrated in Figure \ref{fig:visibilityShortest}, when flying towards the target viewpoint, the drone reaches a position where it cannot frame both actors. When using of the solution  based on the Toric space, it ensures this constraint; the drone continuously maintains the visibility over both actors during the whole flight (see Figure \ref{fig:visibilityCinema}).

\begin{figure}
\begin{subfigure}{\linewidth}
\includegraphics[width=\linewidth]{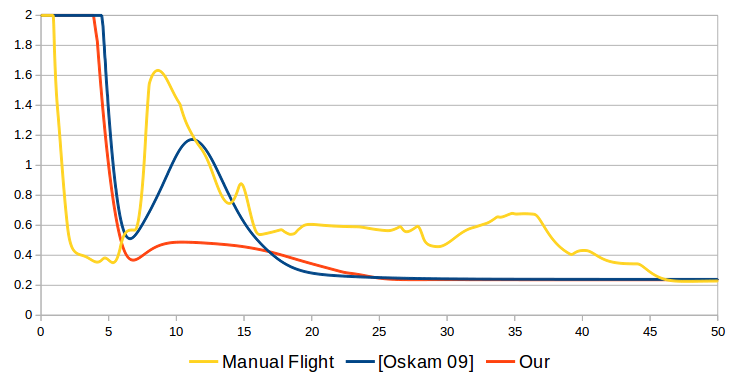}
\caption{Sreen position}
\label{fig:screenPositionDiff}
\end{subfigure}
\begin{subfigure}{\linewidth}
\includegraphics[width=\linewidth]{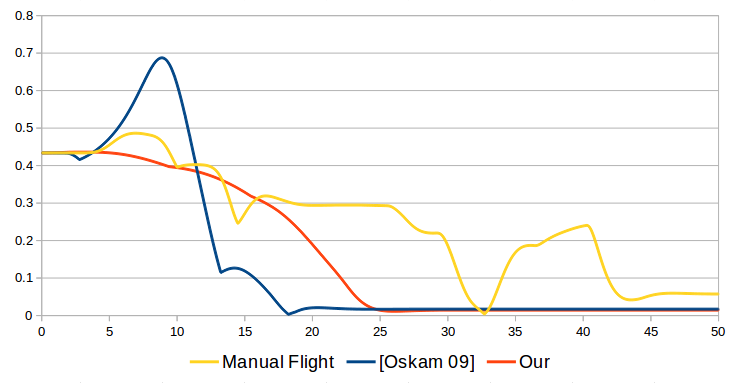}
\caption{Scale}
\label{fig:screenScaleDiff}
\end{subfigure}
\begin{subfigure}{\linewidth}
\includegraphics[width=\linewidth]{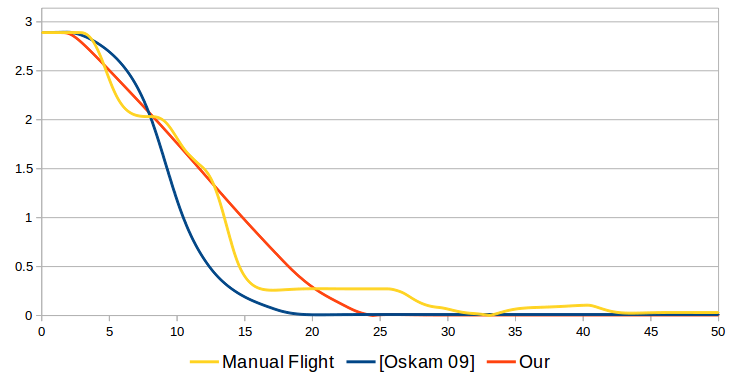}
\caption{Orientation}
\label{fig:orientationDiff}
\end{subfigure}

\caption{The evolution of on screen properties along a drone path. The distance (from 0 when properties values match to 2 when they are of opposite value) between the current framing and the desired framing is displayed along three features: screen position (a), screen size (b) and view angle (c). For each feature, values are computed as the average distance over both actors.}
\end{figure}

The interest of our Drone Toric Space is highlighted in Figures \ref{fig:screenPositionDiff} to \ref{fig:orientationDiff} which show the evolution of visual properties along time. We here analyze the onscreen positions, sizes and view angles of actors -- more precisely, for each feature, we have computed the distance between their desired and actual values. 
 
As previously observed, visibility issues can also be noticed in Figure \ref{fig:screenPositionDiff}. While our cinematographic approach can ensure a small error in terms of onscreen position, \cite{oskam2009visibility} (\ie a shortest path) fails as the camera gets too close to the actors. The same observation can be made on the variation of actors sizes (Figure \ref{fig:screenScaleDiff}). Conversely, our solution provides a smooth interpolation from the initial to the target viewpoint whereas the shortest path suffers from important variations of the actors sizes. 
In a way similar, for the view angle on actors (Figure \ref{fig:orientationDiff}), our solution produces an almost linear interpolation between the initial and target viewpoints. In the case of \cite{oskam2009visibility} this variations is very slow at the beginning, then exponentially increases, and finally stabilizes to its target value.
Footage from these experiments is shown in the companion video.


\subsection{Performance}
\label{performance}

{\bf Path planning.}
Avoiding dynamic obstacles requires the ability to perform path planning tasks in reactive time. To demonstrate the performance of our system and its capacity to avoid dynamic objects, we tested it on different scene configurations. Due to the limitation of the real environment and to provide a more extensive benchmark, we conducted part of this evaluation with the simulator -- which has no impact on the performances of our computations.
Performances of our path planning process are presented in Table \ref{tab:performances}. These results clearly show that all computations time remains lower than 200ms.
\begin{table*}
\centering
	\begin{tabular}{| l || c | c | c | c |}
	\hline
	\textbf{Environment resolution} & 12m x 18m x 4 & 18m x 18m x 4m & 18m x 18m x 4m & 30m x 30m x 5m\\
	\hline
	\textbf{\# spheres} in the roadmap & 1222 & 2320 & 1979 & 2551\\
	\hline
	\textbf{\# nodes} in the roadmap & 8020 & 17247 & 13873 & 22446\\
	\hline
	\textbf{Scene complexity} & medium & low & high & high \\
	\hline
	\hline
	\textbf{Path length} in framing mode (\# nodes) & 14 & 9.7 & 13.4 & 8.1\\
	\hline
	\textbf{Planning duratio}n in framing mode (ms) & 55 & 67 & 69 & 64 \\
	\hline
	\hline
	\textbf{Path length} in sketching mode (\# nodes) & 16.1 & 36.8 & 31.2 & 43.4 \\
	\hline
	\textbf{Planning duration} in sketching mode (ms)& 98 & 87 & 124 & 107\\
	\hline
	\hline
	\textbf{Path optimization duration} (ms) & 76 & 58 & 59 & 78 \\
	\hline
	\end{tabular}
	\caption{Computational performances of our two path planning techniques (framing and sketching).}
	\label{tab:performances}
\end{table*}

{\bf Multi-drone.}
Given the low number of drones used in practice (2 to 4), the computational cost of the min-conflict process remains tractable. The number of combinations for two targets is $n_d^{|\mathcal{F}|}$ where $n_d$ represents the number of drones, and $|\mathcal{F}|$ the number of framings. For $n_d=3$ there are 4913 configurations, and for $n_d=4$ there are 83521 configurations. Table~\ref{tab:conflicts} presents the computation times for distance, visibility and angle conflict detection as well as for tentative path planning (to evaluate the cost of moving to a tentative framing instance), when computing an initial assignment (smoothing is not performed).

 \begin{table}[htbp!]
    \begin{center}
     
    \caption{\label{tab:conflicts} Computation times in ms for the min-conflict optimization in the initial assignment stage for 2 and 4 drones. The last column represents the average time per frame (in ms) of the dynamic local repair reassignement.}
    \begin{tabular}{|l|r|r|r|}
      \hline
      Process & 2 drones (ms) & 4 drones (ms) & local repair\\
      \hline
      Distance 		  	& 1   & 3   & 1\\
      Angle 		  	& 1   & 4   & 1\\
      Visibility 	  	& 47  & 130 & 4\\
      Planning 		  	& 30  & 460 & 3 \\
      Min-conflict 	& 92  & 980 & -\\      
      \hline
    \end{tabular}
    
    \end{center}

 \end{table}

\begin{figure}[t!]
	\centering
		\includegraphics[width=\linewidth]{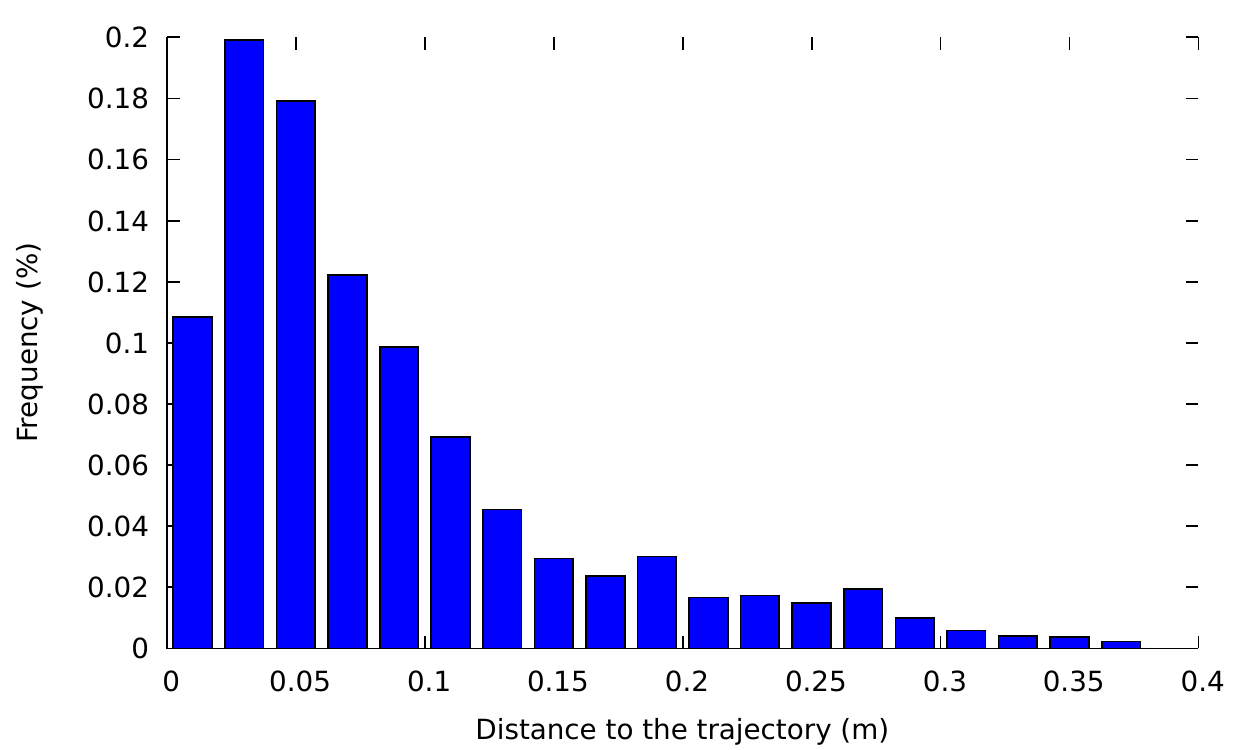}
	\caption{Error in meters on the drone's position with regards to a generated trajectory to follow.}
	\label{fig:distanceError}
\end{figure}

{\bf Precision.}
Figure \ref{fig:distanceError} illustrates the average distance (i) between the drone trajectory to follow and the drone location along time, then (ii) between the desired framing and the actual framing. These results were taken from all the experiments we have conducted with the drones. It demonstrates the precision in world space, and proves the ability of our system to both produce a feasible trajectory and move a drone to closely follow it.


\subsection{Limitations}
The system is currently limited in several aspects. 
First, from a purely hardware standpoint, the choice of the Parrot ARDrone clearly impaired our system capacities. While this drone remain one of the best options in terms of users' safety, the incapacity to tilt the camera, the absence of any form of stabilization and the noisy propellers make it a poor choice for cinematographic tasks.
In future work, we plan to integrate the more recent Parrot Bebop drone. Less noisy this new drone is equipped with a full HD wide angle camera that allows to perform digital tilt and stabilization. Much smaller, it also creates less air perturbations. 
Regarding the planning process, while our solution is fast enough to allow constant recomputations of the path, framing higly moving targets remains challenging. Investigating anticipation schemes appears as a good lead for more advanced path planning strategies.
Finally, our path following method could be improved. As we let the user manually control the drone's velocity along the trajectory, we could further improve the flight accuracy by integrating the local optimization scheme proposed by \cite{Naegeli2017}. Though we were not aware of their work when writing this paper, their optimization scheme is complementary to our path planning technique.

%% file: conclusion.tex
We presented a system to intuitively control one or more cinematographic drones in dynamic scenes. Our system empowers quadrotor drones with cinematographic knowledge, to then enable the design and execution of quadrotor shots, as well as the automated coordination of autonomous drone cinematographers covering a set of moving targets.
Through this paper, we have introduced a model dedicated to the control of drones which ensures the feasibility of drone positions and the safety of targets. We have proposed a complete real-time computation pipeline that enables generating and interactively executing feasible drone trajectories.
We have presented a cinematographic through-the-lens control method adapted to the specificity of controlling quadrotor drones in real environments. We finally proposed an automated technique to orchestrate, in real-time, the simultaneous placement of multiple drones to follow dynamic targets in a cinematographically-sound manner. To the best of our knowledge, this is the first system to provide both interactive and automated cinematographic control on one or multiple quadrotor drones with dynamic targets. We also feel that this kind of systems allows to envision great perspectives towards prototyping and creation of cinematographic sequences, where users would essentially focus on aesthetic choices.